\documentclass[preprint,12pt]{elsarticle}

\usepackage{amssymb}
\usepackage{amsmath}
\usepackage{mathtools}
\usepackage{booktabs}
\usepackage{graphicx}
\usepackage{wrapfig}
\usepackage{caption}
\usepackage{subfigure}
\usepackage{algorithm}
\usepackage{algorithmic}
\usepackage{hyperref}
\usepackage[dvipsnames]{xcolor}
\usepackage{pifont}
\usepackage{xspace}
\usepackage{multirow}
\usepackage{makecell}
\usepackage{enumitem}
\usepackage{csvsimple}
\usepackage{xstring}
\usepackage{silence}
\usepackage{bbm}
\usepackage{microtype}
\usepackage[section]{placeins}

\usepackage{todonotes}

\journal{International Journal of Forecasting}
\usepackage{amsthm}
\newtheorem{theorem}{Theorem}
\newtheorem{lemma}{Lemma}
\newtheorem{proposition}{Proposition}
\newtheorem{corollary}{Corollary} 
\theoremstyle{definition}
\newtheorem{assumption}{Assumption}
\newtheorem{remark}{Remark}
\theoremstyle{plain}

\newcommand{\dataset}[1]{\texttt{#1}\xspace}
\newcommand{\Mone}{\dataset{M1}}
\newcommand{\Mfour}{\dataset{M4}}

\newcommand{\Mthree}{\dataset{M3}}

\newcommand{\Tourism}{\dataset{Tourism}}

\newcommand{\model}[1]{\texttt{#1}\xspace}

\newcommand{\Naive}{\model{Naive}}

\newcommand{\ESRNN}{\model{ESRNN}}
\newcommand{\NHITS}{\model{NHITS}}
\newcommand{\NBEATS}{\model{NBEATS}}

\newcommand{\MQForecaster}{\model{MQForecaster}}
\newcommand{\DeepAR}{\model{DeepAR}}

\newcommand{\DAM}{\model{DAM}}

\newcommand{\LSTM}{\model{LSTM}}
\newcommand{\CNN}{\model{CNN}}
\newcommand{\Transformer}{\model{Transformer}}
\newcommand{\Sfour}{\model{S4}}

\newcommand{\ETS}{\model{ETS}}
\newcommand{\THeta}{\model{Theta}}
\newcommand{\ARIMA}{\model{ARIMA}}

\newcommand{\MS}{MS\xspace}
\newcommand{\REC}{REC\xspace}
\newcommand{\DIR}{DIR\xspace}

\newcommand{\ours}{CH\xspace}
\newcommand{\ourscomplete}{Continuous Horizon }

\providecommand{\cmark}{\textcolor{teal}{\ding{51}}}%
\providecommand{\xmark}{\textcolor{red}{\ding{55}}}%

\newcommand{\SE}[1]{\scriptsize \scalebox{.7}{({\color{gray}{#1}})} }

\newcommand\bw{\boldsymbol{w}}
\newcommand\bphi{\boldsymbol{\phi}}
\newcommand\btheta{\boldsymbol{\theta}}

\DeclareMathOperator*{\argmin}{arg\,min}

\begin{document}

\begin{frontmatter}

\title{Continuous Horizon Forecasting}

\affiliation[inst1]{organization={Forecasting Science, Amazon},
            city={New York},
            country={NY}}

\affiliation[inst2]{organization={University of California},
            city={Berkeley},
            country={CA}}

\affiliation[inst3]{organization={Mohamed bin Zayed University of Artificial Intelligence},
            country={UAE}}

\author[inst1]{Malcolm Wolff\texorpdfstring{\corref{cor1}}{}}
\ead{wolfmalc@amazon.com}
\cortext[cor1]{Corresponding Author}
\author[inst1]{Elliott Zackrone}
\author[inst3]{Souhaib Ben Taieb}
\author[inst1,inst2]{\\Michael W. Mahoney}
\author[inst1]{Kin G. Olivares}


\begin{abstract}
Multi-horizon forecasting has been dominated by discrete forecast strategies that treat model outputs as fixed-vectors or recurrently create them. In this paper, we introduce the \ourscomplete (\ours) strategy that generalizes upon recursive, direct, and multi-step strategies by redefining forecasts as a continuous operator and the forecast horizon as a query variable rather than an index. We present the theoretical foundations for the \ours strategy, a Universal Operator Approximation Theorem, and a continuous-horizon bias-variance decomposition that formally proves the estimation variance improvements of \ours over the discrete strategies. We validate the \ours strategy empirically through a simulation study, and on 16 large-scale datasets from the M-series benchmark, showing how \ours achieves competitive RMSE accuracy, while uniquely enabling parallel inference, targeted horizon querying, and interpolation and extrapolation across unseen horizons.
\end{abstract}

\end{frontmatter}

\section{Introduction} \label{sec:01_introduction}

Multi-horizon forecasting has been dominated by discrete strategies: multi-step (\MS;~\citep{taieb2010multiple_output, bao2014msvr}), direct (\DIR;~\citep{chevillon2007direct}) and recurrent (\REC)~\citep{taieb2012review}.
Although tree-based machine learning models~\citep{januschowski2022trees}, neural architectures~\citep{oreshkin2020nbeats, challu2023nhits, nie2023patchtst}, and foundation models~\citep{garza2023timegpt, google2024timesfm, aws2024chronos, salesforce2023moirai, hoo2024tabpfn} have advanced rapidly, multi-horizon forecast strategies have remained fixed for nearly a decade: multi-step fixed-grid outputs or recurrent rollouts~\citep{salinas2020deepar, smyl2019esrnn}, with their known trade-offs in error accumulation, output dimension, and inability to query arbitrary horizons~\citep{taieb2015bias}.

\begin{table*}[t]
\centering
\caption{Desirable properties of forecasting strategies.}
\label{tab:comparison}
\scriptsize
\begin{tabular}{@{}lcccc@{}}
\toprule
Property & Recursive & Direct & Multi-Step & \ourscomplete \\
\midrule
Low Bias                              & \xmark & \cmark & \cmark & \cmark \\
Low Variance & \xmark & \xmark & \xmark & \cmark \\
Inference parallelism                    & \xmark & \cmark & \cmark & \cmark \\
Targeted horizon querying     & \xmark & \cmark & \xmark & \cmark \\
Unseen Horizon inter/extrapolation & \xmark & \xmark & \xmark & \cmark \\
Joint horizon modeling    & \xmark & \xmark & \cmark & \cmark \\
\bottomrule
\end{tabular}
\end{table*}

We argue that this discrete view of multi-horizon strategies is a historical artifact rather than a necessity. Time series forecasting has long recognized future trajectories as inherently continuous phenomena \citep{ramsay2005functional,hyndman2007functional_data_approach, chapados2007functional_time_series, hyndman2009functional_time_series}. In this paper, we elevate this continuous data approach into a forecast strategy: \ourscomplete (\ours). We reveal that discrete strategies are coarse discretizations of \ours's continuous-horizon forecast operator, we unlock theoretical understanding and showcase the strategy's empirical advantages:

\begin{enumerate}[label=(\roman*), itemsep=.5pt]
    \item \textbf{Universal Approximation \& Unseen Horizons}: We theoretically and empirically show that the \ourscomplete (\ours) strategy can approximate any continuous function across arbitrary horizons, including fractional horizons and those unseen during training.

    \item \textbf{Continuous Bias-Variance Dominance}: We extend the discrete MSE decomposition framework of \citet{taieb2015bias} to continuous time, formally proving how \ours leverages continuous effective horizon information to reduce estimation variance over traditional multi-step (\MS) and direct (\DIR) strategies.

    \item \textbf{Accuracy, Computational Efficiency \& Parallelism}: \ours achieves competitive accuracy results on the M-series benchmark (16 large-scale datasets), while enabling flexible inference that supports both targeted single-horizon queries and batched parallel inference.
\end{enumerate}

The remainder of this paper is structured as follows. 
Section~\ref{sec:02_notation} introduces the multi-horizon forecasting notation. 
Section~\ref{sec:03_theory} presents the theoretical foundations of the \ours strategy. 
Section~\ref{sec:04_theory_validation} conducts simulation of the strategies' bias-variance behavior. Section~\ref{sec:05_interp_extrap} investigates the strategies' horizon interpolation and extrapolation capabilities. Section~\ref{sec:06_evaluation} presents the empirical results of the M-series and inference computational complexity analysis. 
Section~\ref{sec:07_discussion} hosts the discussion and Section~\ref{sec:08_conclusion} concludes.
 \clearpage
\begin{figure}[t] 
    \centering
    \subfigure[\emph{Recursive}]{\label{fig:recurrent}
    \includegraphics[width=0.45\linewidth]{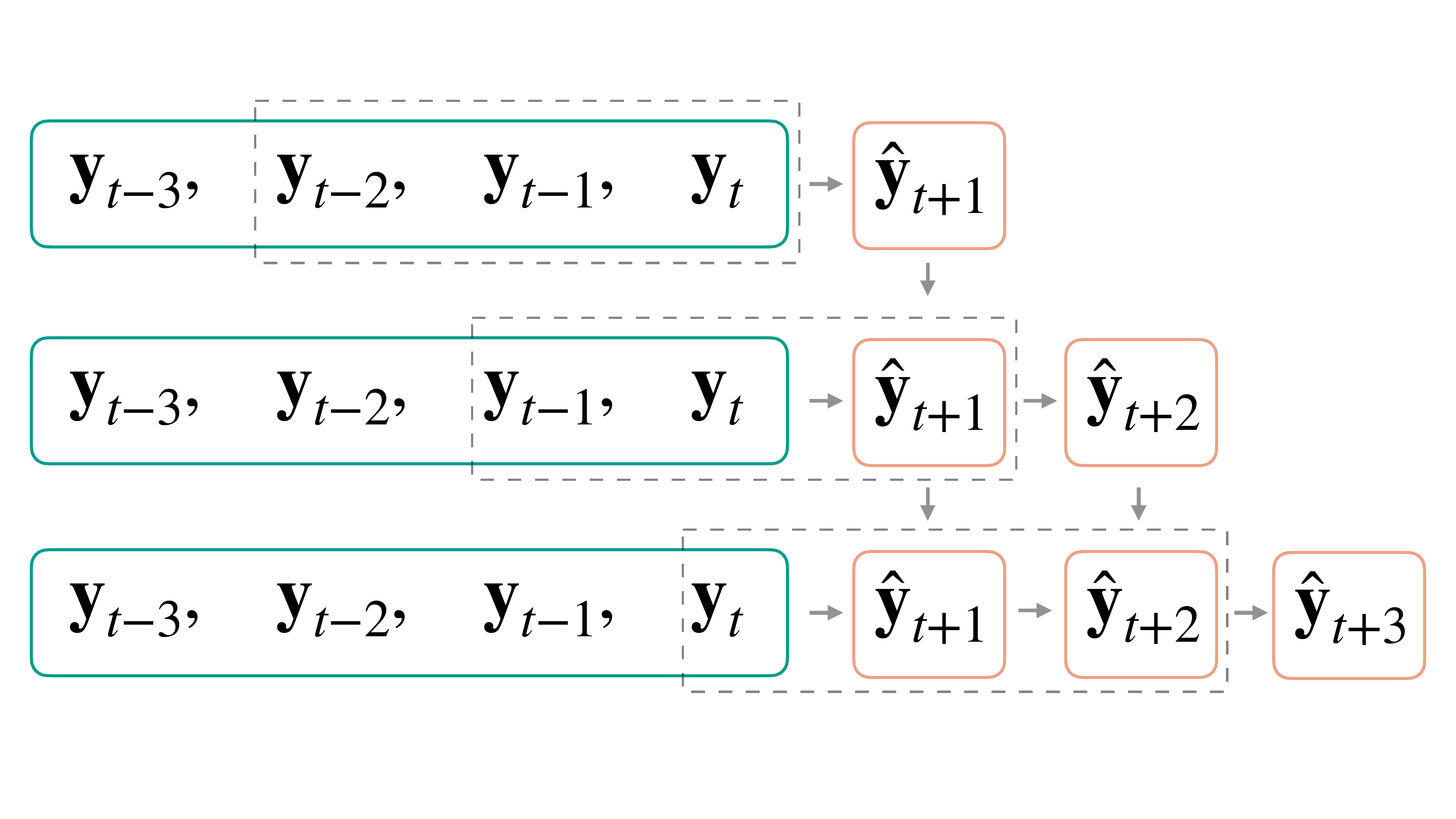}} 
    \subfigure[\emph{Multi-Step}]{\label{fig:multistep}
    \includegraphics[width=0.45\linewidth]{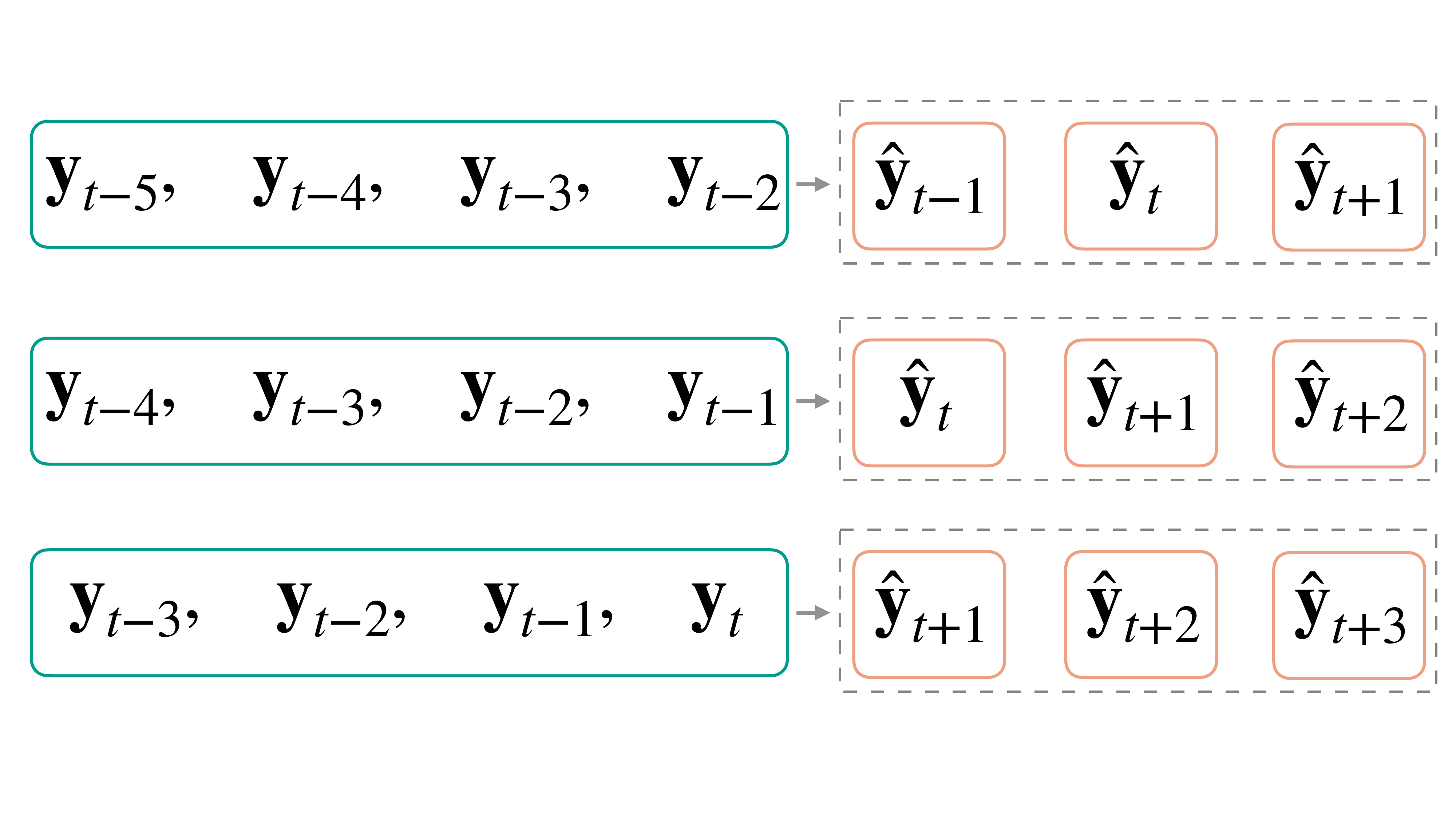}} 
    \subfigure[\emph{Direct}]{\label{fig:direct}
    \includegraphics[width=0.45\linewidth]{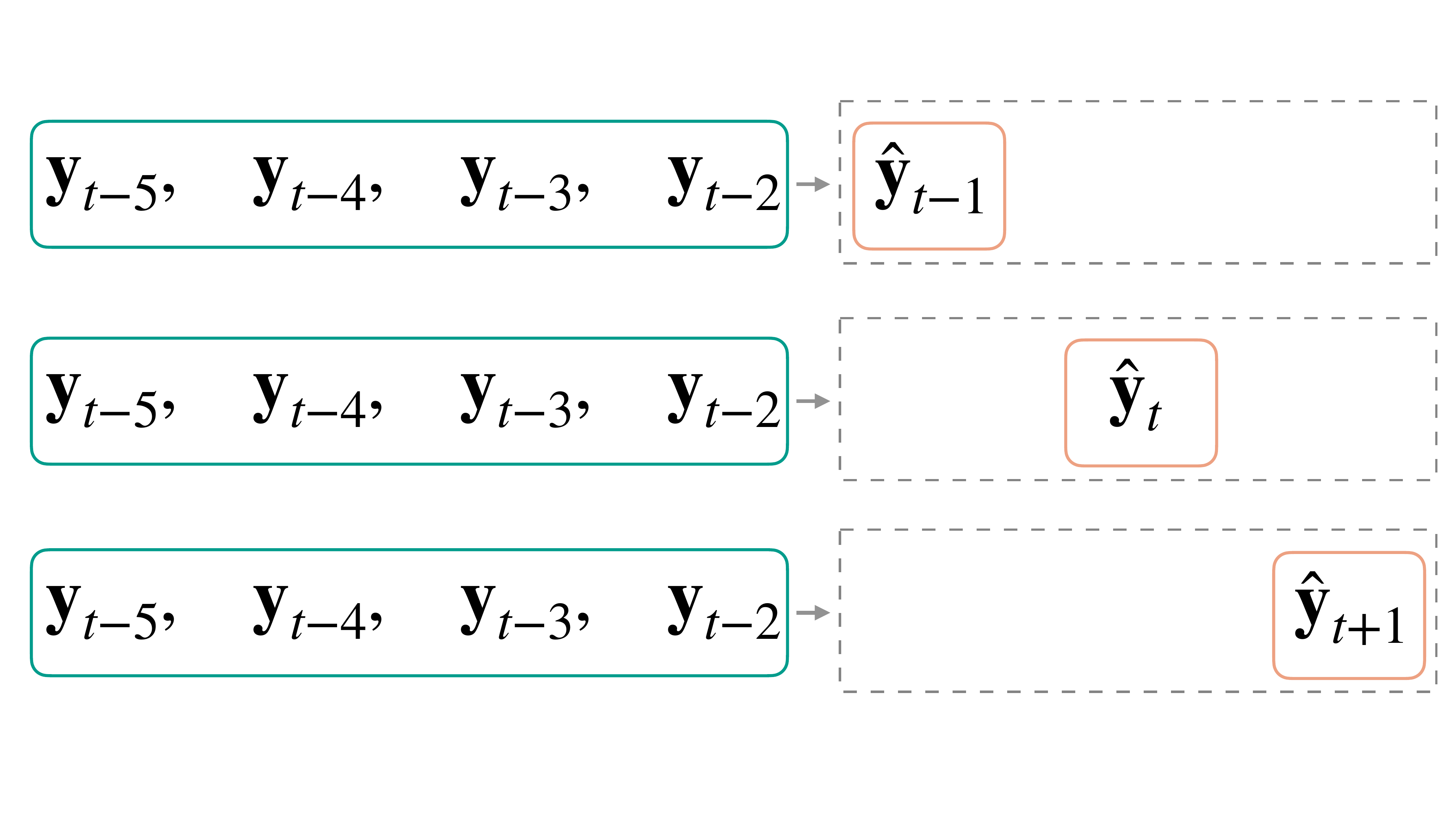}} 
    \subfigure[\emph{\ourscomplete}]{\label{fig:ours}
    \includegraphics[width=0.45\linewidth]{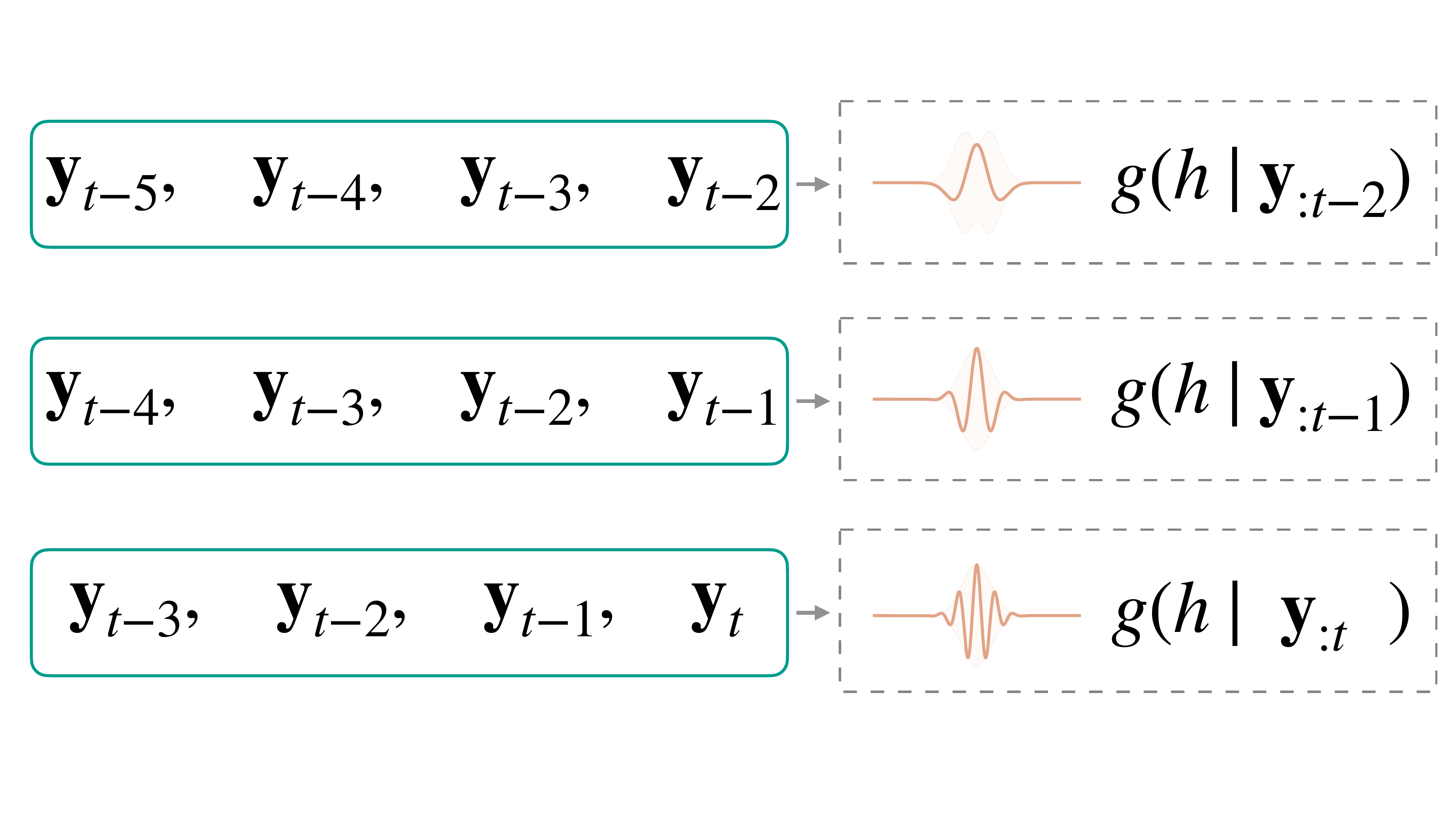}} 
    \caption{
    Multi-horizon forecast strategies and the continuous-horizon paradigm.
    }
\label{fig:forecast_strategies}
\end{figure}

\section{Multi-Horizon Forecasting}
\label{sec:02_notation}

Here we introduce general multi-horizon forecasting task notation. Let the forecast creation date be denoted by the index $t$.  We denote the functional random variable as $Y$, we denote its historical discrete observations at FCD $t$ as $\mathbf{y}_{:t}=[y_1, y_{2}, \ldots, y_t]$, for a single horizon $h$ the variable of interest is the future behavior of $Y_{t+h}$, for $H$ horizons $\mathbf{Y}_{t+1:t+H}=[Y_{t+1}, Y_{t+2}, \ldots, Y_{t+H}] \in \mathbb{R}^{H}$. 

Following the functional time series view ~\citep{hyndman2007functional_data_approach, ramsay2005functional}, we assume that the target variable comes from a continuous process:
\begin{equation}
    Y_{t+h} = f(h \mid \mathbf{y}_{:t}) + \epsilon_{t+h}, \qquad \mathrm{where} \quad f(\cdot \mid \mathbf{y}_{:t}) \in \mathcal{C}([0,H])
\end{equation}

with $f$ a continuous data generating process (DGP), and $\epsilon_{t+h}$ is a zero-mean noise term. The DGP coincides with the conditional expectation, $\mathbb{E}[Y_{t+h} \mid \mathbf{y}_{:t}]=f(h \mid \mathbf{y}_{:t}) = \mu_{t+h|t}$.

We denote the $h$-step ahead forecast $\hat{y}_{t+h} = g(h \,\mid\,
\mathbf{y}_{:t})$, where $g$ is the model. We denote vector outputs $\mathbf{\hat{y}}_{t+1:t+H}=[\hat{y}_{t+1}, \hat{y}_{t+2}, \ldots, \hat{y}_{t+H}] \in \mathbb{R}^{H}$. The model's input handling and the structure of its forecast output $\mathbf{\hat{y}}_{t+1:t+H}$ define the overarching multi-horizon forecasting strategy.

\subsection{Multi-Horizon Forecast Strategies}

Inspired by the taxonomy of \citet{taieb2012review}, we distinguish the primary variants of the multi-horizon forecast strategy: \emph{recursive}, \emph{direct}, \emph{multi-step} and \emph{continuous horizon}. 

\begin{itemize}
    \item \textbf{Recursive (\REC):} A single one-step-ahead model, that is, iterated H-times, by feeding each prediction back as input for the next step. It is a simple and parameter efficient approach that suffers from error accumulation.
    \begin{equation}
        \hat{y}_{t+h} = g\left(1 \,\mid\, \mathbf{y}_{:t},\, \hat{y}_{t+1},\, \ldots,\, \hat{y}_{t+h-1} \right), \qquad h = 1,\dots,H.
    \end{equation}

    \item  \textbf{Direct (DIR):} H separate models are trained, one per horizon. Each model directly predicts the value of step h without dependencies on intermediate predictions. It avoids error accumulation, but assumes conditional independence and is less statistically efficient.
    \begin{equation}
        \hat{\mathbf{y}}_{t+1:t+H} = [g_{1}\left(\mathbf{y}_{:t}  \right),\dots,g_{H}\left(\mathbf{y}_{:t}\right)].
    \end{equation}

    \item \textbf{Multi-Step (MS):} Also known as \emph{multiple-input multiple output}. A single model outputs all H-forecasts simultaneously as a vector. It preserves the stochastic dependencies between horizons, as it does not assume conditional independence.
    \begin{equation}
        \hat{\mathbf{y}}_{t+1:t+H} = g\left(\mathbf{y}_{:t}\right) \in \mathbb{R}^{H}.
    \end{equation}

    \item \textbf{\ourscomplete (\ours):} A single model produces the forecast on horizon $h$ through a continuous function conditioned on $h$.
    \begin{equation}
    \begin{aligned}
        \hat{y}_{t+h} &= g(h \mid \mathbf{y}_{:t}), \quad g(\cdot \mid \mathbf{y}_{:t}) \in \mathcal{C}([0,H]), \\
        g(h \mid \mathbf{y}_{:t}) &= \sum_{k=1}^{K} \btheta_{k}(\mathbf{y}_{:t})\, \bphi_{k}(h).
    \end{aligned}
    \label{eq:ch_definition}
    \end{equation}
\end{itemize}


The \ourscomplete strategy generalizes the other three: recursive, direct, and multi-step forecasts are all \emph{discretizations} of a continuous horizon operator, obtained by evaluating an operator at fixed integer grid points or by iterating a one-step map. Redefining the forecast from a discrete mapping to an operator that returns a forecast function, that can be thought as the asymptotic limit of a forecast horizon grid. In this work we use a flexible bilinear parametrization of \ours, similar to that of the \citet{hyndman2007functional_data_approach} and the branch-trunk architectures~\citep{lu2021deep_onet}.

 \clearpage
\section{Theoretical Foundations} 
\label{sec:03_theory}

\subsection{Parameter Learning}
Estimation of a forecasting model $g$ is carried out according to 
\begin{equation}
    \hat{\mathbf{w}} := \argmin_{\mathbf{w}}\;
    \frac{1}{T}\sum_{t=1}^{T}
    \int_{0}^{H}
    L\big(y_{t+h},\, g(h \mid \mathbf{y}_{:t}, \mathbf{w})\big)\,\mathrm{d}\rho(h).
    \label{eq:erm}
\end{equation}
where the forecaster parameters $\mathbf{w}$ are selected to minimize loss $L$. Strategies can be identified by a way to select horizons on which to train. We can describe a common learning objective using a horizon-selector measure $\rho$ on the horizon interval $[0, H]$ that specifies the lead times entering the objective: a point mass $\delta_h$ for the \REC and \DIR strategies, the empirical measure $\tfrac{1}{H}\sum_{j=1}^H \delta_{h_j}$ for \MS, and the uniform measure $\tfrac{1}{H}\,\mathrm{d}h$ for the \ours strategy.

\subsection{Mean Squared Error Decomposition Analysis}

A deep understanding of the performance of forecasting strategies can be gained through the analysis of the mean squared error decomposition into bias and variance components.

\begin{lemma}[MSE decomposition]
\label{lem:taieb_mse}
The forecast MSE at horizon $h$ can be decomposed as
\begin{equation}
\begin{aligned}
\mathrm{MSE}_h
&=
\mathbb E_{\mathbf{y}_{:t}}\!\Big[
\underbrace{\mathbb E_{\varepsilon,\mathcal D}\!\big[(y_{t+h}- g(h\mid \mathbf{y}_{:t}, \hat{\bw}))^2 \mid \mathbf{y}_{:t}\big]}_{\mathrm{MSE}_h(\mathbf{y}_{:t})}
\Big]
\\[4pt]
&=
\mathbb E_{\mathbf{y}_{:t}}\!\Big[
\underbrace{\mathbb E_{\varepsilon}\!\big[(y_{t+h}-\mu_{t+h|t})^2 \mid \mathbf{y}_{:t}\big]}_{N_h(\mathbf{y}_{:t})}
+
\underbrace{\big(\mu_{t+h|t}-\mathbb E_{\mathcal D}[g(h\mid \mathbf{y}_{:t}, \hat{\bw})]\big)^2}_{B_h(\mathbf{y}_{:t})}
\\[2pt]
&\qquad\quad
+
\underbrace{\mathbb E_{\mathcal D}\!\big[(g(h\mid \mathbf{y}_{:t}, \hat{\bw})-\mathbb E_{\mathcal D}[g(h\mid \mathbf{y}_{:t}, \hat{\bw})])^2 \mid \mathbf{y}_{:t}\big]}_{V_h(\mathbf{y}_{:t})}
\Big].
\end{aligned}
\label{eq:taieb14}
\end{equation}
This decomposition separates the error at each horizon into an irreducible noise term $N_h$, a squared bias term $B_h$, and an estimation variance term $V_h$. The noise $N_h$ is the conditional variance of $y_{t+h}$ around the optimal forecast $\mu_{t+h|t}$; it is irreducible, set by the DGP rather than the model, and typically grows with the horizon. The squared bias $B_h$ measures how far the average fitted forecast $\mathbb E_{\mathcal D}[g(h\mid\mathbf{y}_{:t},\,\hat{\bw})]$ sits from $\mu_{t+h|t}$, and reflects the model's lack of flexibility to represent the target variable. The estimation variance $V_h$ measures the sensitivity of the fitted forecast to the particular training samples, and shrinks as the effective number of observations informing $\hat{\bw}$ grows.
\end{lemma}

\begin{lemma}[Bias decomposition]
\label{lem:taieb_bias}
The squared bias decomposes as
\begin{equation}
\begin{aligned}
B_h
&=
\mathbb E_{\mathbf{y}_{:t}}\!\Big[\big(\mu_{t+h|t}-\mathbb E_{\mathcal D}[g(h\mid \mathbf{y}_{:t}, \hat{\bw})]\big)^2\Big]
\\[4pt]
&=
\mathbb E_{\mathbf{y}_{:t}}\!\left[
\Big(
\underbrace{\big(\mu_{t+h|t}-g(h\mid \mathbf{y}_{:t},\bw^\star)\big)}_{\text{approximation}}
+
\underbrace{\big(g(h\mid \mathbf{y}_{:t},\bw^\star)-\mathbb E_{\mathcal D}[g(h\mid \mathbf{y}_{:t}, \hat{\bw})]\big)}_{\text{estimation}}
\Big)^2
\right] .
\end{aligned}
\label{eq:bias_decomposition}
\end{equation}
The approximation term measures the discrepancy between the optimal forecast $\mu_{t+h|t}$ and the best forecast attainable by the model, independent of any estimation of a finite-sample; the estimation term measures how the training-average estimator departs from that best-in-class forecast under a finite sample.
\end{lemma}

Model selection trades bias $B_h$ against variance $V_h$; the two results below show how the Continuous Horizon strategy controls each in turn.

\begin{theorem}[Universal Operator Approximation Theorem] Let $\mathcal{Y} \subset \mathbb{R}^{t}$ be a compact input domain of historical observations $\mathbf{y}_{:t} \in \mathcal{Y}$ and $[0,H]$ be a compact forecast horizon interval. Let the DGP process of the target variable be governed by a continuous operator such as $F: \mathcal{Y} \to \mathcal{C}([0,H])$, with $F(\mathbf{y}_{:t})=f(\cdot\mid\mathbf{y}_{:t})$. Then for any accuracy $\epsilon > 0$, there exists a neural network approximation $g\left(h \mid \mathbf{y}_{:t}, \bw^{\star} \right)=\sum^{K}_{k=1} 
    \btheta_{k}\left(\mathbf{y}_{:t}\right) \cdot \bphi_{k}(h)$, composed of a finite set of continuous neural basis functions $\{\bphi_{k}(h)\}^{K}_{k=1}$ and coefficient mappings $\{\btheta_{k}(\mathbf{y}_{:t})\}^{K}_{k=1}$, such that:
\begin{equation}
\begin{split}
    \int_{0}^{H} \left( f\left(h \mid \mathbf{y}_{:t}\right) - g\left(h \mid \mathbf{y}_{:t}, \bw^{\star}\right) \right)^2 dh
    \qquad\qquad\qquad \qquad\qquad\quad\\
    = \int_{0}^{H} \left( f\left(h \mid \mathbf{y}_{:t}\right) - \sum^{K}_{k=1} 
    \btheta_{k}\left(\mathbf{y}_{:t}\right) \cdot \bphi_{k}(h) \right)^2 dh < \epsilon, \quad \forall \mathbf{y}_{:t} \in \mathcal{Y}.
\end{split}
\end{equation}
\label{theorem:universal_approximation}
\end{theorem}
\vspace{-.5cm}
Theorem~\ref{theorem:universal_approximation} resembles the operator approximation results of \citet{chen1995approximation}, which also support the DeepONet~\citep{lu2021deep_onet} and Fourier neural operators~\citep{li2021fourierneuraloperatorparametric}, and the neural basis approximation theorem of \NHITS~\citep{challu2023nhits}. We include Theorem~\ref{theorem:universal_approximation} to show that the separable bilinear decoder in Eq.~\eqref{eq:ch_definition} loses no expressiveness and to bound the approximation term of $B_h$ (Lemma~\ref{lem:taieb_bias}). Proof available in \ref{appendix:universal_approximation}.

\begin{theorem}[Continuous Horizon Variance Dominance]
\label{thm:ch_dominates_body}
Consider residuals $(y_{t+h}-\hat{y}_{t+h})$ and let $K(h, h') \in [0, 1]$ be a stationary, non-increasing, strictly convex kernel that describes the residual covariance at horizons $h,h'$, so that $K(h,h')=K(|h-h'|)$. For the multi-step discrete forecast grid $[1,2,\dots,H] \subset [0, H]$, we define the effective horizon information ratio as:
\begin{equation}
\mathcal{I}_{\mathrm{eff}}^{\mathrm{MS}} = \frac{H^2}{\sum_{h=1}^H \sum_{h'=1}^H K(h, h')} \quad \text{and} \quad \mathcal{I}_{\mathrm{eff}}^{\mathrm{CH}} = \frac{H^2}{\int_0^H \int_0^H K(h, h') \, \mathrm{d}h \mathrm{d}h'}.
\end{equation}
Then, the effective horizon information of the \ours, \MS, and \DIR strategies obeys the inequality:
\begin{equation}
1 = \mathcal{I}_{\mathrm{eff}}^{\mathrm{DIR}} < \mathcal{I}_{\mathrm{eff}}^{\mathrm{MS}} < \mathcal{I}_{\mathrm{eff}}^{\mathrm{CH}}.
\end{equation}
Consequently, for a matched capacity of shared parameters $P$, the estimation variance satisfies the following:
\begin{equation}
V_{\mathrm{CH}}(h) < V_{\mathrm{MS}}(h) < V_{\mathrm{DIR}}(h), \quad \forall h \in [0, H],
\end{equation}
\label{theorem:variance_dominance}
\end{theorem}

The proof of the variance dominance theorem is available in \ref{appendix:ch_decomposition}.

\begin{figure}[t] 
    \centering
    \subfigure[\emph{Effective Horizon Information}]{\label{fig:information}
    \includegraphics[width=0.45\linewidth]{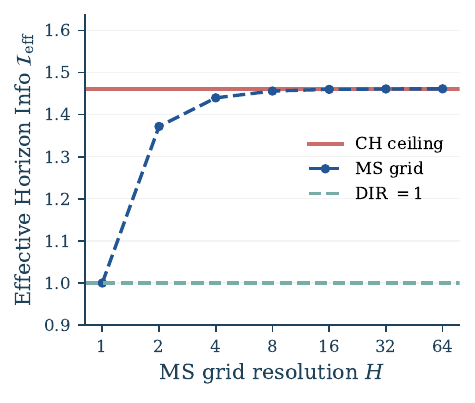}} 
    \subfigure[\emph{Estimation Variance}]{\label{fig:variance}
    \includegraphics[width=0.45\linewidth]{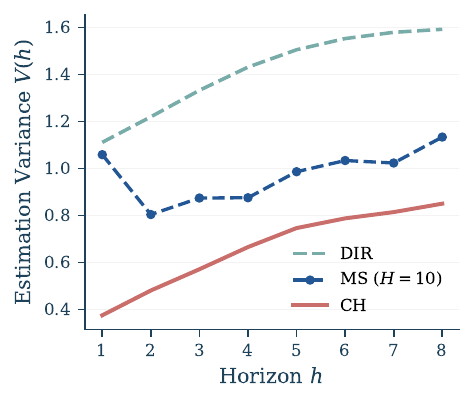}} 
    \caption{
    Effective information and estimation variance for different \MS resolution and forecast horizon. The estimation variance of \ours dominates both \MS and \DIR.
    }
\label{fig:thm2_synthetic1}
\end{figure}

\clearpage
\section{Bias and Variance Analysis by Simulation}
\label{sec:04_theory_validation}

\begin{figure*}[t!]
    \centering
    \subfigure[\emph{LAR MSE decomposition}]{\label{fig:thm2_lar}
    \includegraphics[width=0.48\linewidth]{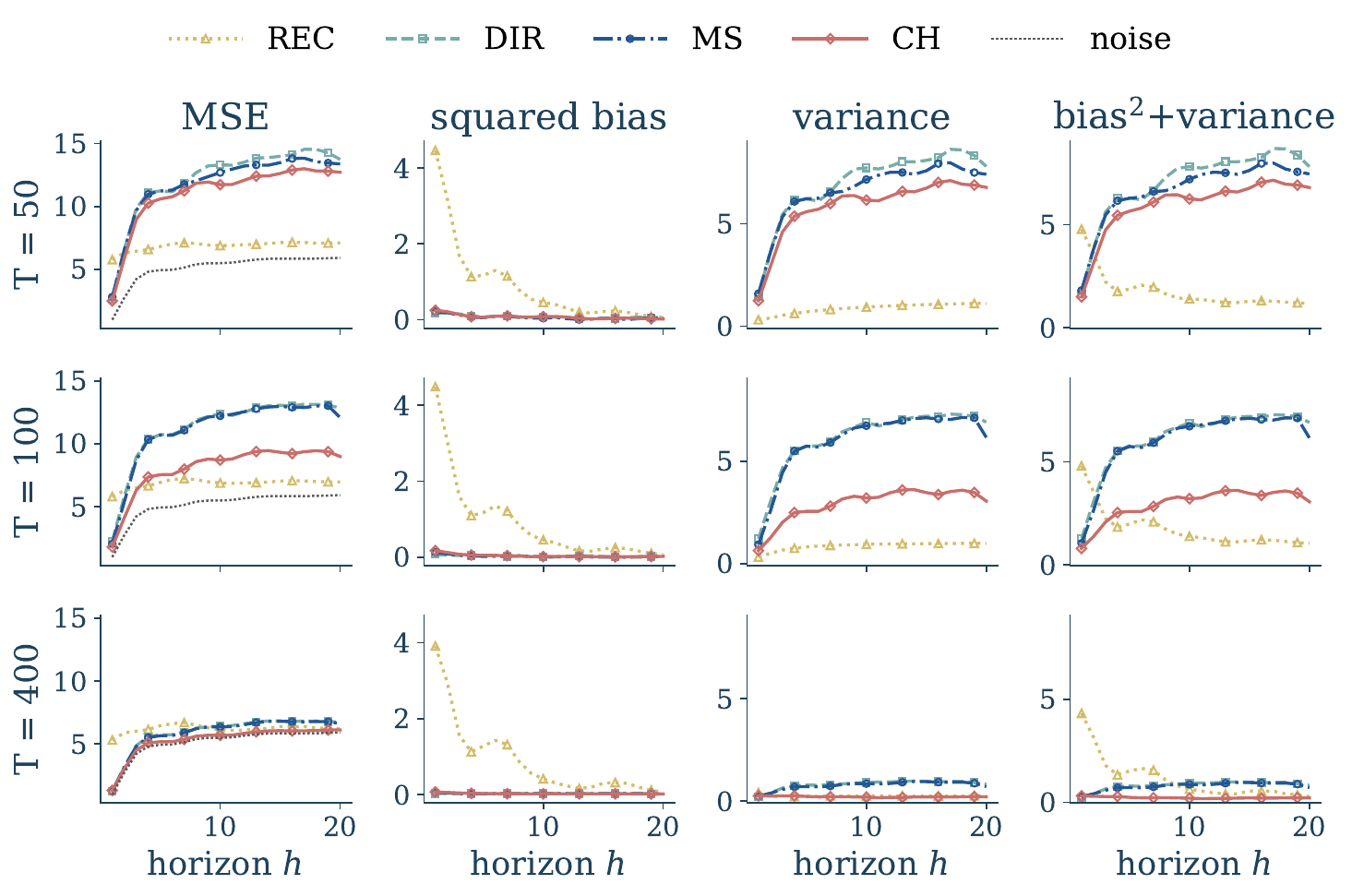}}
    \subfigure[\emph{Bias--variance versus predictability}]{\label{fig:thm2_noise}
    \includegraphics[width=0.48\linewidth]{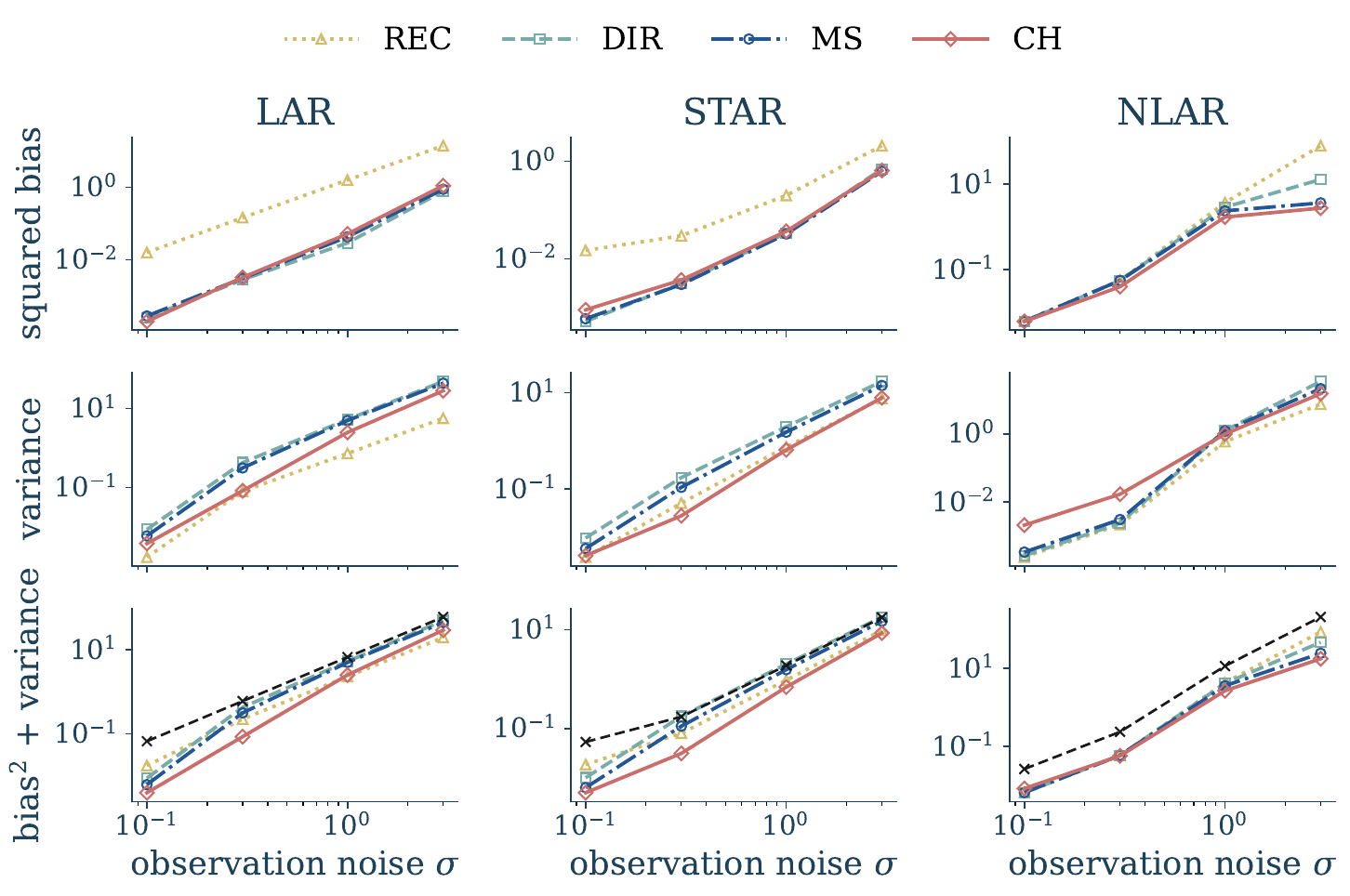}}
    \caption{(a) MSE decomposition for a LAR DGP. (b) Bias--variance components across noise levels for NLAR, STAR, and LAR processes. \ours\ improves the bias--variance tradeoff compared to \MS and \DIR strategies, an advantage that grows in higher-noise regimes. The dashed Naive-1 reference (bias$^2$+variance row) bounds the useful noise range: the learned strategies stay below it where the process is predictable; its collapse toward the irreducible noise floor marks the loss of predictability.
    }
\label{fig:thm2_synthetic2}
\end{figure*}

To validate Theorem~\ref{theorem:variance_dominance}, we do two simulation studies.

\subsection{Controlled Basis Experiment.}

We replace the learned neural components with a fixed basis of polynomials and harmonics, and we use a Gaussian process as the DGP. Thus, the experiment measures only the estimation variance. Figure~\ref{fig:thm2_synthetic1} shows that, with the matched parameters, \ours\ is confirmed to have a lower estimation variance $V(h)$ than \MS\ and \DIR\ on all horizons.

\subsection{Classical Autoregressive DGPs.}

The second study uses the DGPs and the Monte Carlo protocol of \citet{taieb2015bias}. We use an MQForecaster~\citep{wen2017recurrentforecaster} with a \ours\ decoder. The model must learn both the coefficient mappings and the basis functions $\hat{\phi}_k(h)$. We use three autoregressive processes: a linear autoregression (LAR;~\citealp{box2015time}), a non-linear autoregression (NLAR;~\citealp{medeiros2006building, kock2011forecasting}), and a smooth-transition autoregression (STAR;~\citealp{fan2005nonlinear, terasvirta2010modelling}). \ref{appendix:dpgs_details} gives the details of the DGPs.

Figure~\ref{fig:thm2_lar} shows the MSE decomposition of the LAR process for different training lengths $T$. On this linear process, \MS, \DIR, and \ours\ have a bias near zero. \DIR\ has a high estimation variance, because it fits independent parameters for each horizon. \MS\ decreases the variance, because it shares parameters across a discrete grid of horizons. \ours\ has the lowest estimation variance and the lowest total error at all horizons $h$. \ref{appendix:nlar_decomposition} and \ref{appendix:star_decomposition} repeat the bias-variance analysis for NLAR and STAR.

Figure~\ref{fig:thm2_noise} shows the bias--variance components when we increase the observation noise, which is a proxy for a weaker correlation between horizons. When noise increases, the estimation variance of \DIR\ and \MS\ increases rapidly. The shared continuous basis of \ours\ regularizes the fit across horizons. At the highest noise level, \ours\ has approximately 40 to 55\% lower variance than \DIR\ on all three DGPs. \ours\ also has the lowest total error on NLAR and STAR, but on LAR, \REC\ has a lower total error. %

\begin{figure}[t]
    \centering
    \includegraphics[width=\linewidth]{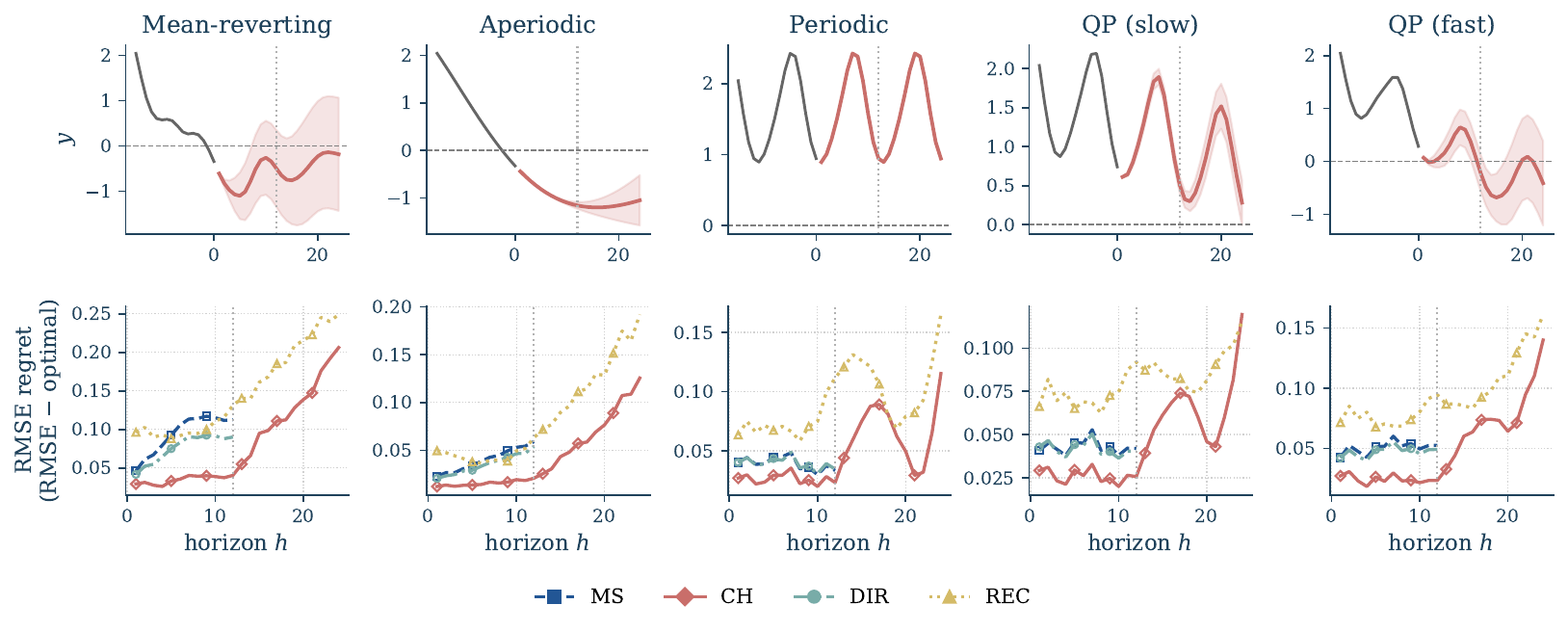}
    \caption{
    Interpolation and extrapolation on five Gaussian processes. \textbf{Top:} A sampled history (gray) and the exact posterior mean with its $10$--$90\%$ band. The dotted line shows the end of the training grid ($h=12$).
    \textbf{Bottom:} The RMSE regret at each horizon (Eq.~\eqref{eq:gp_regret}). A regret of zero is optimal. \ours has the lowest regret on almost all seen and unseen horizons. \MS and \DIR are shown only on their training grid.
    }
    \label{fig:gp_regimes_combined}
\end{figure}

\section{Interpolation and Extrapolation to Unseen Horizons}
\label{sec:05_interp_extrap}

Discrete strategies such as \MS and \DIR give forecasts only at fixed horizons. \ours can query its basis functions $\bphi_k(h)$ at any real $h$. Thus, \ours can interpolate at fractional horizons, for example $h=\pi$, and it can extrapolate to unseen horizons. Of the discrete strategies, only the recurrent rollout of \REC can also extrapolate. To show these capabilities, we simulate continuous-time Gaussian processes in five persistence regimes: mean-reverting, aperiodic, periodic, and slow and fast quasi-periodic (Figure~\ref{fig:gp_regimes_combined}, \ref{appendix:interpolation_extrapolation}).

We train the MQForecaster model on the integer horizons $h \le 12$. We evaluate it on three sets of horizons: the seen integers, the unseen half-integers (interpolation), and the unseen integers up to $h=24$ (extrapolation). For the half-integer horizons, \MS and \DIR use linear interpolation between their two adjacent output horizons. We measure the accuracy with the RMSE and with the regret relative to the exact analytical posterior.
\begin{equation}
\mathrm{RMSE}(\mathbf{y}, \mathbf{\hat{y}}) = \sqrt{\frac{1}{T \times |\mathcal{H}|} \sum_{t=1}^{T}\sum_{h \in \mathcal{H}} \left(y_{t+h}-\hat{y}_{t+h}\right)^{2}},
\label{eq:rmse}
\end{equation}
\begin{equation}
\mathrm{Regret}(\mathbf{y}, \mathbf{\hat{y}}, \boldsymbol{\mu}) = \mathrm{RMSE}(\mathbf{y}, \mathbf{\hat{y}}) - \mathrm{RMSE}(\mathbf{y}, \mu_{t+h|t}),
\label{eq:gp_regret}
\end{equation}

where $\mathcal{H}$ is a set of uniformly spaced query horizons, and $\mu_{t+h|t} = f(h \mid \mathbf{y}_{:t})$ is the exact conditional mean of Section~\ref{sec:02_notation}. For the seen and the interpolation sets, $\mathcal{H} \subset [0,H]$. For the extrapolation set, $\mathcal{H} \subset (H, 2H]$.

Figure~\ref{fig:gp_regimes_combined} (bottom) and Table~\ref{table:interp_extrap} show that \ours has the lowest error on all five processes and on all three sets of horizons. Under interpolation, \ours has a lower error than the interpolated \MS and \DIR forecasts. \ref{appendix:qualitative_forecasts} explains this result as manifestation of aliasing of a discrete horizon grid.

In extrapolation, \ours reverts to the unconditional mean on the aperiodic processes and follows the cyclic dynamics with small ripples. The \REC rollout has a higher error than \ours on all five processes. For periodic targets, Theorem~\ref{theorem:periodic_approximation} (\ref{appendix:extrapolation_bounds}) extends the approximation guarantee beyond the training horizons $[0, H]$.

\begin{table*}[t]
\centering
\caption{
Interpolation and extrapolation on five Gaussian-process regimes (RMSE, lower is better). All strategies train on the horizons $h\le12$. We evaluate on the seen integers, the unseen half-integers (interpolation), and the unseen integers up to $h=24$ (extrapolation). For the half-integers, \MS and \DIR use linear interpolation between their output horizons. \MS and \DIR have no output for $h>12$, but \ours and \REC give forecasts for these horizons. The best value in each block is \textbf{bold}.
}
\label{table:interp_extrap}
\scriptsize
\setlength{\tabcolsep}{4pt}
\begin{tabular}{@{}l cccc @{\quad} cccc @{\quad} cc@{}}
\toprule
 & \multicolumn{4}{c}{Seen ($h\le12$)} & \multicolumn{4}{c}{Interpolation} & \multicolumn{2}{c}{Extrapolation} \\
\cmidrule(lr){2-5}\cmidrule(lr){6-9}\cmidrule(l){10-11}
Process & \ours & \MS & \DIR & \REC & \ours & \MS & \DIR & \REC & \ours & \REC \\
\midrule
Mean-reverting & \textbf{0.873} & 0.934 & 0.915 & 0.939 & \textbf{0.872} & 0.932 & 0.908 & 0.937 & \textbf{1.211} & 1.280 \\
Aperiodic      & \textbf{0.733} & 0.759 & 0.754 & 0.763 & \textbf{0.730} & 0.753 & 0.746 & 0.758 & \textbf{1.054} & 1.110 \\
Periodic       & \textbf{0.568} & 0.581 & 0.581 & 0.617 & \textbf{0.568} & 0.576 & 0.575 & 0.613 & \textbf{0.607} & 0.649 \\
QP (slow)      & \textbf{0.623} & 0.640 & 0.639 & 0.671 & \textbf{0.623} & 0.636 & 0.633 & 0.670 & \textbf{0.756} & 0.780 \\
QP (fast)      & \textbf{0.689} & 0.716 & 0.712 & 0.743 & \textbf{0.688} & 0.712 & 0.705 & 0.740 & \textbf{0.940} & 0.975 \\
\bottomrule
\end{tabular}
\end{table*}

 \clearpage
\section{Empirical Evaluation} \label{sec:06_evaluation}

\begin{figure*}[t!]
    \centering
    \includegraphics[width=.99\textwidth]{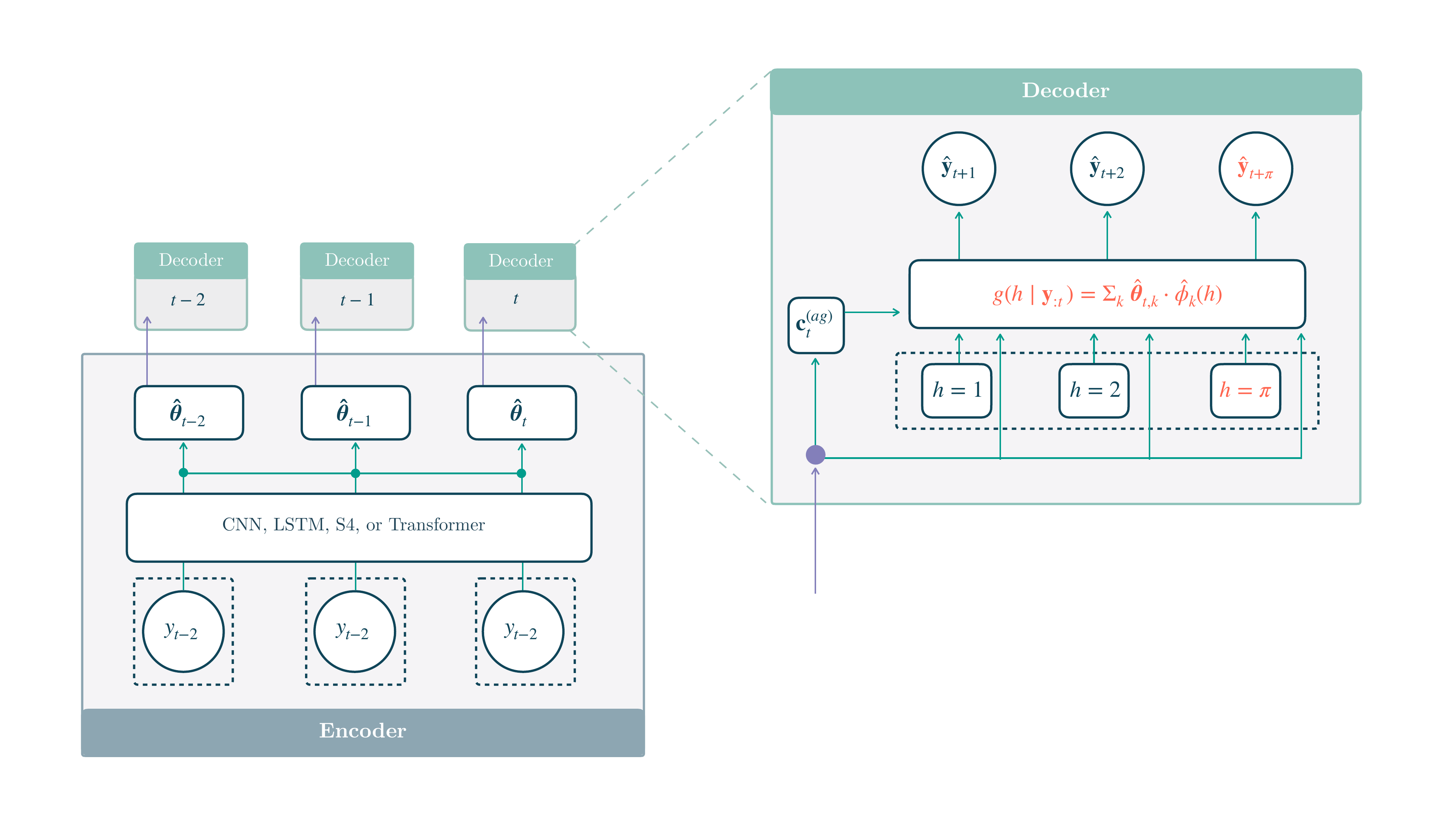}
    \caption{
    \textbf{Continuous Horizon (CH) decoder within the \MQForecaster architecture.} 
    Instead of outputting a discrete vector, the CH forecast strategy outputs a function $g(h \mid \mathbf{y}_{:t})$, capable of evaluating continuous horizons $h \in [0, H]$ on demand. This enables direct evaluation at arbitrary continuous lead times (including unseen interpolated and extrapolated steps) while maintaining parameter sharing for lower estimation variance. Holding the backbone fixed isolates the strategy, confirming that the \ours advantages generalize across encoders.
    }
\label{fig:ch_decoder}
\end{figure*}

\subsection{Neural Forecast Architecture}  \label{sec:06_architecture}

We evaluate all strategies within the unified \MQForecaster architecture~\citep{wen2017recurrentforecaster,olivares2021pmm_cnn} sequence-to-sequence model~\citep{cho2014Seq2SeqC}. Its modular encoder-decoder design provides a natural testbed to isolate the impact of the forecasting strategy while holding architecture constant (Figure~\ref{fig:ch_decoder}). Historical realizations $\mathbf{y}_{:t}$ are encoded into a hidden state $\btheta_k(\mathbf{y}_{:t})$.

\begin{equation}
     \btheta_k(\mathbf{y}_{:t}) = \mathrm{Encoder}\left(\mathbf{y}_{:t}\right)
\end{equation}

A decoder then maps the hidden state to the forecast. We keep the encoder fixed and change only the decoder to realize each forecast strategy.

The strategies \MS and \DIR instantiate the decoder exactly as their taxonomy definitions (Section~\ref{sec:02_notation}) prescribe, replacing the abstract function $g$ with an MLP acting on the hidden state; therefore, we omit their repetition here. For the \REC strategy, we employ a similar approach to that of the \DeepAR decoding strategy~\citep{salinas2020deepar}.

To enable the \ours strategy, we use a bilinear expansion approximator informed by the universal approximation theorem (Theorem~\ref{theorem:universal_approximation}):
\label{sec:continuous_decoder}
\begin{equation}
    g(h\mid\mathbf{y}_{:t}) = \sum^{K}_{k=0} \btheta_k(\mathbf{y}_{:t}) \cdot \bphi_{k}(h), \qquad \bphi_{0}(h) = 1,
    \label{eq:ch_concat}
\end{equation}
where the constant basis function $\hat{\bphi}_{0}=1$ gives a level $\hat{\btheta}_{t,0}$ that is the same for all horizons. The horizon basis functions are a learned embedding table; see the ablation study in \ref{appendix:basis_ablation}, where we explore basis variants. Based on this ablation, we use tabular embedding basis functions ($K=16$). \ref{appendix:ch_objective} describes the controlled learning objective of the M-series experiments.

\begin{figure}[t!]
    \centering
    \includegraphics[width=0.65\linewidth]{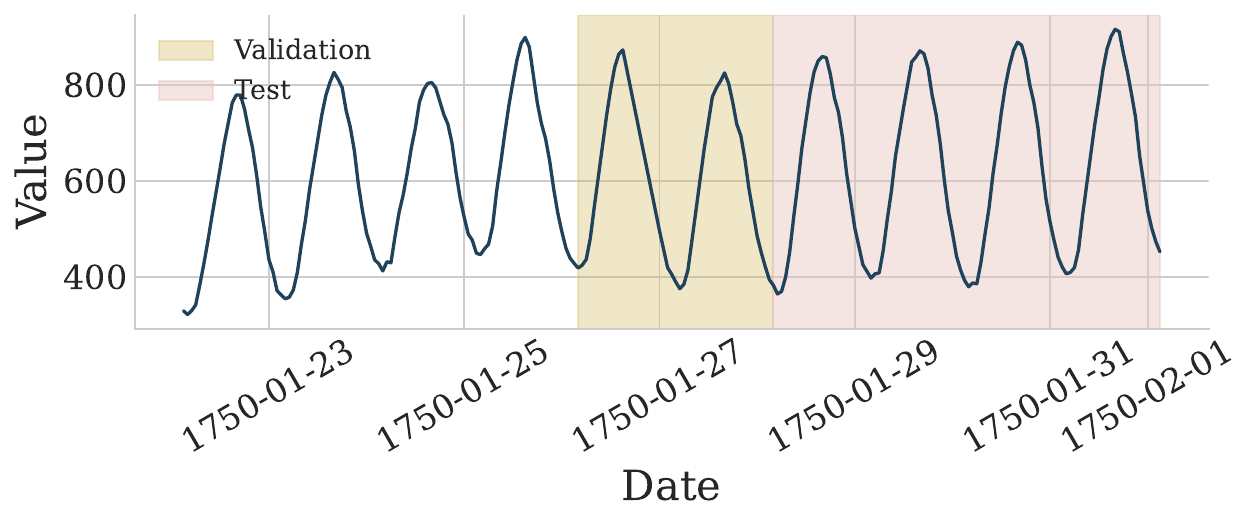}
    \vspace{-0.2cm}
    \caption{
    Temporal cross-validation on an \Mfour\ hourly series, with training (blue), validation (yellow, $H$), and test (salmon, $2H$) regions.
    }
    \label{fig:training_methodology}
\end{figure}

\subsection{Benchmark Datasets} 
\label{sec:benchmark_datasets}

We evaluate our approach on the M-series benchmark, comprising 16 large-scale datasets with more than 100,000 series from the competitions \Mone~\citep{makridakis1982m1_competition}, \Mthree~\citep{makridakis2000m3_competition}, \Mfour~\citep{makridakis2020m4_competition}, and \Tourism~\citep{athanosopoulus2011tourism_competition}. Following recent literature~\citep{olivares2025cftl_evaluation, potosnak2026forkingsequences}, we use temporal cross-validation by holding out the final $3H$ observations per series (Details in \ref{appendix:dataset_details}): the first $H$ form the validation window, and the final $2H$ comprise the test window (Figure~\ref{fig:training_methodology}). Rolling the test FCD forward yields $H$ overlapping evaluations covering $[T-2H+1, T]$. Denoting the final series index by $T$, the data partitions are defined as:
\begin{equation}
\begin{split}
\mathbf{y}_{\text{train}} = \mathbf{y}_{\::\: T-3H}, \quad
\mathbf{y}_{\text{validation}} = \mathbf{y}_{T-3H+1 \::\: T-2H}, \quad
\mathbf{y}_{\text{test}} = \mathbf{y}_{T-2H+1 \::\: T}.
\end{split}
\end{equation}

\subsection{Main Empirical Results}

\begin{table}[t!]
\caption{Empirical evaluation of point forecasts. Relative Average RMSE (relAvgRMSE) with the \Naive baseline, averaged over 5 runs. Lower measurements are preferred; values below 1 beat the \Naive. The methods without standard deviation have deterministic solutions. For each encoder we compare four forecasting strategies: our \ourscomplete (\ours), multi-step (MS), direct (\DIR), and recurrent (\REC).}

\label{table:main_mse}
\centering
\scriptsize
\resizebox{\textwidth}{!}{
\begin{tabular}{ll|cccc|cccc|cccc|cccc|cc}
\toprule
& & \multicolumn{4}{c|}{\LSTM}
& \multicolumn{4}{c|}{\CNN}
& \multicolumn{4}{c|}{\Transformer}
& \multicolumn{4}{c|}{\Sfour}
& \multicolumn{2}{c}{StatsForecast} \\
& Freq 
& \multicolumn{1}{c}{\ours} & \multicolumn{1}{c}{MS} & \multicolumn{1}{c}{\DIR} & \multicolumn{1}{c|}{\REC} 
& \multicolumn{1}{c}{\ours} & \multicolumn{1}{c}{MS} & \multicolumn{1}{c}{\DIR} & \multicolumn{1}{c|}{\REC}
& \multicolumn{1}{c}{\ours} & \multicolumn{1}{c}{MS} & \multicolumn{1}{c}{\DIR} & \multicolumn{1}{c|}{\REC}
& \multicolumn{1}{c}{\ours} & \multicolumn{1}{c}{MS} & \multicolumn{1}{c}{\DIR} & \multicolumn{1}{c|}{\REC} 
& \ETS & \ARIMA \\
\midrule
\xdef\prevdataset{}
\begin{filecontents*}{external_paper/tables/mse_main_table.csv}
Dataset,Frequency,LSTM_OURS,LSTM_OURS_SE,LSTM_MS,LSTM_MS_SE,LSTM_DR,LSTM_DR_SE,LSTM_RC,LSTM_RC_SE,CNN_OURS,CNN_OURS_SE,CNN_MS,CNN_MS_SE,CNN_DR,CNN_DR_SE,CNN_RC,CNN_RC_SE,Transformer_OURS,Transformer_OURS_SE,Transformer_MS,Transformer_MS_SE,Transformer_DR,Transformer_DR_SE,Transformer_RC,Transformer_RC_SE,S4_OURS,S4_OURS_SE,S4_MS,S4_MS_SE,S4_DR,S4_DR_SE,S4_RC,S4_RC_SE,ETS,ARIMA
M1,M,0.8098,0.0060,0.8784,0.0061,0.9244,0.0048,0.9281,0.0149,0.8226,0.0050,0.9023,0.0078,0.9150,0.0222,0.9928,0.0417,0.9022,0.0226,0.9373,0.0095,0.9715,0.0215,0.9741,0.0082,0.8705,0.0120,0.8781,0.0078,0.9366,0.0277,0.9891,0.0159,0.7347,0.7625
M1,Q,0.9305,0.0264,0.9441,0.0207,0.9616,0.0314,0.9172,0.0268,0.9185,0.0053,0.9359,0.0179,0.9969,0.0244,1.2225,0.1147,1.0004,0.0809,1.0238,0.0178,1.0329,0.0515,0.9454,0.0251,0.9160,0.0214,0.9031,0.0047,1.0435,0.0234,0.9264,0.0235,0.8998,0.9071
M1,Y,0.8426,0.0245,0.7564,0.0207,0.7958,0.0240,0.9285,0.0805,0.8014,0.0122,0.8323,0.0289,0.7367,0.0473,0.7786,0.0353,0.6896,0.0267,0.8307,0.0770,0.8448,0.0720,0.8032,0.0530,0.7384,0.0328,0.7950,0.0281,0.7750,0.0434,0.8660,0.0559,0.7854,0.7696
M3,O,0.2348,0.0020,0.2429,0.0214,0.2414,0.0069,0.2651,0.0159,0.1844,0.0012,0.2192,0.0033,0.2422,0.0078,0.3834,0.0712,0.2507,0.0121,0.2798,0.0099,0.4130,0.1321,0.2930,0.0233,0.1870,0.0030,0.1979,0.0040,0.2568,0.0153,0.3134,0.0200,0.7146,0.7259
M3,M,0.6973,0.0042,0.7134,0.0071,0.7807,0.0062,0.8662,0.0261,0.7222,0.0011,0.7445,0.0028,0.7465,0.0036,1.1237,0.0447,0.7618,0.0169,0.7904,0.0098,0.8213,0.0168,1.0170,0.0674,0.7115,0.0034,0.7353,0.0059,0.7503,0.0046,0.9843,0.0255,0.8380,0.8490
M3,Q,0.7962,0.0095,0.8061,0.0111,0.8685,0.0113,0.9332,0.0224,0.7849,0.0074,0.8035,0.0069,0.8541,0.0126,1.0807,0.0890,0.8780,0.0481,0.9300,0.0333,0.9650,0.0137,0.9288,0.0454,0.7561,0.0111,0.7390,0.0038,0.9114,0.0040,0.9702,0.0142,0.9206,0.9279
M3,Y,0.8879,0.0130,0.7854,0.0104,0.8758,0.0108,0.9264,0.0458,0.8230,0.0100,0.7939,0.0084,0.7772,0.0152,0.9710,0.0678,0.8685,0.0674,0.8596,0.0428,0.8605,0.0236,0.8324,0.0236,0.7996,0.0240,0.7793,0.0133,0.8778,0.0114,0.9198,0.0113,0.9678,1.0116
M4,H,0.1719,0.0016,0.1810,0.0056,0.3916,0.0519,0.6788,0.1059,0.2850,0.0097,0.4610,0.0058,0.4739,0.0333,0.7736,0.0892,0.2471,0.0157,0.4877,0.0721,0.6908,0.0369,0.9976,0.0652,0.2698,0.0030,0.3016,0.0235,0.2773,0.0311,0.7000,0.0755,0.4158,0.1987
M4,D,0.8362,0.0071,0.7923,0.0122,0.8536,0.0183,0.9311,0.0347,0.8344,0.0097,0.8293,0.0067,0.8531,0.0118,0.9065,0.0540,0.8687,0.0270,0.8401,0.0190,0.8717,0.0203,1.0212,0.0660,0.8178,0.0128,0.8068,0.0098,0.8458,0.0173,1.0401,0.0491,0.9815,0.9879
M4,W,0.7839,0.0203,0.7240,0.0076,0.8100,0.0089,0.8016,0.0107,0.7662,0.0027,0.7894,0.0042,0.7887,0.0155,0.8604,0.0274,0.8286,0.0134,0.8333,0.0189,0.8363,0.0155,0.8828,0.0219,0.8055,0.0105,0.7943,0.0080,0.8271,0.0210,0.9093,0.0337,0.9507,0.9032
M4,M,0.7114,0.0029,0.7307,0.0127,0.7482,0.0081,0.8337,0.0181,0.7278,0.0040,0.7305,0.0040,0.7322,0.0073,1.0405,0.0623,0.7198,0.0114,0.7490,0.0102,0.7841,0.0389,1.0134,0.0716,0.7054,0.0055,0.7343,0.0079,0.7297,0.0091,0.9806,0.0358,0.8802,0.8904
M4,Q,0.7426,0.0079,0.7373,0.0083,0.7904,0.0051,0.8072,0.0114,0.7349,0.0015,0.7398,0.0023,0.7520,0.0045,0.8950,0.0257,0.8274,0.0283,0.8158,0.0165,0.8448,0.0138,0.9164,0.0365,0.7580,0.0084,0.7591,0.0077,0.7961,0.0072,0.8868,0.0121,0.8714,0.9143
M4,Y,0.8034,0.0106,0.7394,0.0067,0.8031,0.0063,0.8642,0.0467,0.7515,0.0068,0.7380,0.0054,0.7412,0.0144,0.8591,0.0541,0.7566,0.0550,0.8014,0.0562,0.7970,0.0161,0.7406,0.0237,0.7183,0.0159,0.7162,0.0102,0.8074,0.0059,0.8417,0.0122,0.8758,0.9376
Tourism,M,0.5856,0.0058,0.6011,0.0157,0.7637,0.0271,0.8260,0.0419,0.6533,0.0074,0.6772,0.0085,0.6764,0.0170,1.1139,0.0593,0.8286,0.0071,0.8171,0.0070,0.8317,0.0116,1.0842,0.1034,0.6790,0.0063,0.6217,0.0066,0.6433,0.0075,1.2160,0.0502,0.4476,0.5026
Tourism,Q,0.6702,0.0124,0.6691,0.0140,0.8084,0.0100,0.8133,0.0174,0.6304,0.0063,0.6563,0.0118,0.7421,0.0083,0.9466,0.0314,0.8464,0.0304,0.8733,0.0409,0.8508,0.0071,0.8725,0.0487,0.6389,0.0203,0.6034,0.0182,0.7636,0.0218,0.9239,0.0283,0.6395,0.7091
Tourism,Y,1.0035,0.0351,0.9265,0.0113,0.9803,0.0289,0.9683,0.0216,0.8991,0.0171,0.9517,0.0139,0.9078,0.0219,0.9480,0.0231,0.9138,0.0287,0.9556,0.0456,0.9518,0.0251,0.9431,0.0313,0.9562,0.0337,0.9523,0.0127,1.0954,0.0356,1.0042,0.0392,0.8924,1.0213
\end{filecontents*}
\csvreader[]{external_paper/tables/mse_main_table.csv}
{Dataset=\dset,
Frequency=\frequency,
LSTM_OURS=\lstmours,
LSTM_OURS_SE=\lstmoursse,
LSTM_MS=\lstmms,
LSTM_MS_SE=\lstmmsse,
LSTM_DR=\lstmdr,
LSTM_DR_SE=\lstmdrse,
LSTM_RC=\lstmrc,
LSTM_RC_SE=\lstmrcse,
CNN_OURS=\cnnours,
CNN_OURS_SE=\cnnoursse,
CNN_MS=\cnnms,
CNN_MS_SE=\cnnmsse,
CNN_DR=\cnndr,
CNN_DR_SE=\cnndrse,
CNN_RC=\cnnrc,
CNN_RC_SE=\cnnrcse,
Transformer_OURS=\transours,
Transformer_OURS_SE=\transoursse,
Transformer_MS=\transms,
Transformer_MS_SE=\transmsse,
Transformer_DR=\transdr,
Transformer_DR_SE=\transdrse,
Transformer_RC=\transrc,
Transformer_RC_SE=\transrcse,
S4_OURS=\sfourours,
S4_OURS_SE=\sfouroursse,
S4_MS=\sfourms,
S4_MS_SE=\sfourmsse,
S4_DR=\sfourdr,
S4_DR_SE=\sfourdrse,
S4_RC=\sfourrc,
S4_RC_SE=\sfourrcse,
ETS=\ets,
ARIMA=\arima}
{%
\IfStrEq{\dset}{\prevdataset}{%
    \xdef\prevdataset{\dset}%
    \def\printdataset{}%
}{%
    \xdef\prevdataset{\dset}%
    \def\printdataset{\dset}%
}%

\printdataset & \frequency
& \lstmours & \lstmms & \lstmdr & \lstmrc 
& \cnnours & \cnnms & \cnndr & \cnnrc 
& \transours & \transms & \transdr & \transrc 
& \sfourours & \sfourms & \sfourdr & \sfourrc 
& \ets & \arima \\
& 
& \SE{\lstmoursse} & \SE{\lstmmsse} & \SE{\lstmdrse} & \SE{\lstmrcse} 
& \SE{\cnnoursse} & \SE{\cnnmsse} & \SE{\cnndrse} & \SE{\cnnrcse} 
& \SE{\transoursse} & \SE{\transmsse} & \SE{\transdrse} & \SE{\transrcse} 
& \SE{\sfouroursse} & \SE{\sfourmsse} & \SE{\sfourdrse} & \SE{\sfourrcse}
& \SE{-} & \SE{-} \\
\IfStrEq{\frequency}{Y}{%
    \IfStrEq{\dset}{tourism}{\\\bottomrule}{\\[-4pt]\hline\\[-3pt]}%
}{\hphantom{}}
}
\end{tabular}}
\end{table}

\textbf{Accuracy Metrics.} Following \citet{hyndman2006another_look_measures}, we report the \emph{Relative Average RMSE} (relAvgRMSE), defined as the ratio of the mean per-series RMSE to that of the \Naive\ baseline:
\begin{equation}
\label{equation:relavgrmse}
    \mathrm{relAvgRMSE} =
    \frac{\frac{1}{N}\sum_{i=1}^{N}\mathrm{RMSE}_i(\mathbf{y}^{(i)}, \mathbf{\hat{y}}^{(i)})}
         {\frac{1}{N}\sum_{i=1}^{N}\mathrm{RMSE}_i(\mathbf{y}^{(i)}, \mathbf{\tilde{y}}^{(i)})},
\end{equation}
where $N$ is the number of series and $\mathbf{y}^{(i)}$, $\mathbf{\hat{y}}^{(i)}$, and $\mathbf{\tilde{y}}^{(i)}$ denote the target, model forecast and \Naive\ baseline for series $i$. Computing the root per series before aggregation prevents large-scale series from dominating the metric. Values below $1$ indicate an improvement over \Naive.

\textbf{Forecast Baselines.} We vary \MQForecaster's encoder (Section~\ref{sec:06_architecture}): \CNN~\citep{vandenoord2016wavenet}, \LSTM~\citep{hochreiter1997long, change2017dilatedrnn}, a \Transformer~\citep{vaswani2017attention}, and \Sfour~\citep{gu2022s4}. We include \ETS~\citep{holt1957exponential_smoothing} and \ARIMA~\citep{hyndman2008automatic_arima, hyndman2026forecasting} as baselines. \ref{appendix:hyperparameters} gives the network hyperparameters. \ref{appendix:param_matching} discusses the parameter matching across strategies.

\textbf{Point Forecast Results.} Table~\ref{table:main_mse} compares the strategies at matched parameter counts (\ref{appendix:param_matching}). Thus, a difference in accuracy comes from the forecast strategy, not from the capacity of the model. \ours\ has a lower relAvgRMSE than \MS\ in 40 of the 64 encoder and frequency pairs, with a mean improvement of 2.6\%. In 12 pairs, the improvement is greater than two standard errors and in 8 pairs \MS \ is better by more than two standard errors. The advantage of \ours\ increases with the horizon length, as predicted by Section~\ref{sec:03_theory}. \ours\ is better in all \Mfour\ Hourly pairs ($H=48$), with a decrease of 38\% (\CNN) and 49\% (\Transformer), and in 14 of the 16 monthly pairs. In the short yearly horizons ($H \le 6$), the two strategies are approximately equal, and on \Mfour\ Daily, \MS\ is lower in all four pairs. \ours\ is also better than \REC\ in 58 of the 64 encoder and dataset pairs, with a mean improvement of 19\%.

\ref{appendix:smape_mase} complements the RMSE results with sMAPE and MASE, making our results closer to the M competitions.

\subsection{Inference Complexity and Parallelism}
\label{sec:06_inference_complexity}

\begin{figure}[t]
  \centering
  \begin{minipage}[c]{0.46\textwidth}
    \centering
    \captionof{table}{Decoder cost of one query}
    \label{tab:inference_complexity}
    \scriptsize
    \begin{tabular}{l|cc}
    \toprule
                 & $h \le H$              & $h > H$                      \\ \midrule
    \REC         & $\mathcal{O}(h)$       & $\mathcal{O}(h)$             \\
    \DIR         & $\mathcal{O}(1)$       & $\mathcal{O}(1)^{\dagger}$  \\
    \MS          & $\mathcal{O}(H)$       & $\mathcal{O}(h)^{\dagger}$  \\
    \ours        & $\mathcal{O}(1)$       & $\mathcal{O}(1)$             \\ \bottomrule
    \end{tabular}
  \end{minipage}\hfill
  \begin{minipage}[c]{0.52\textwidth}
    \centering
    \includegraphics[width=\linewidth]{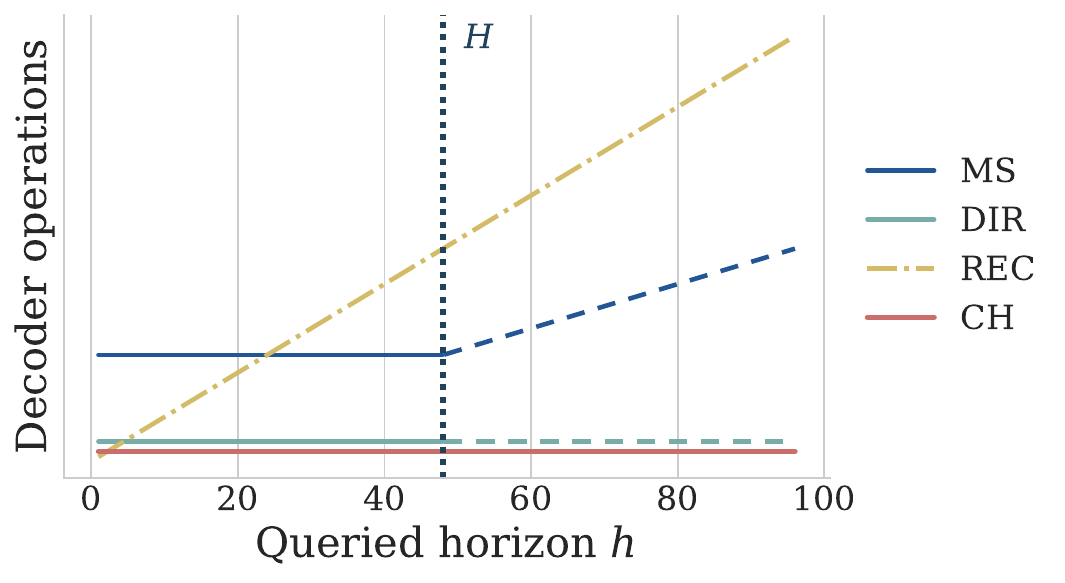}
  \end{minipage}
    \caption{\textbf{Targeted query decoder cost (schematic).} \MS\ forecasts all $H$ horizons. \DIR\ runs horizon-$h$'s model. \REC\ rolls out $h$ sequential steps. \ours\ needs one readout of the cached basis. For $h>H$, \MS\ and \DIR\ need a new model (dashed lines and $^{\dagger}$).}
  \label{fig:inference_cost}
\end{figure}

Beyond accuracy, strategies differ in the decoder cost of one forecast on a queried horizon $h$ (Table~\ref{tab:inference_complexity}, Figure~\ref{fig:inference_cost}). \MS\ computes the full grid of $H$ horizons, so a query costs $\mathcal{O}(H)$. \REC\ rolls out $h$ sequential steps, so one query costs $\mathcal{O}(h)$. \DIR\ runs only the head of horizon $h$, so one query costs $\mathcal{O}(1)$, but \DIR\ must train and store $H$ heads. 

The \ours\ basis functions $\bphi_k(h)$ do not depend on the history, so we cache them before inference. Then a query on any horizon needs only one readout, at a cost of $\mathcal{O}(1)$. For $h>H$, \ours\ uses the same model. When the complete trajectory is necessary, \ours\ evaluates all horizons in parallel, as \MS\ does.
 
\section{Discussion and Related Work} \label{sec:07_discussion}

\ours\ extends the functional time series approach~\citep{ramsay2005functional, hyndman2007functional_data_approach} in three ways. It uses the function directly as an end-to-end forecast strategy, without an in-sample decomposition. It learns the full mapping with a neural network, without smoothing or functional principal components. It learns its basis with MLPs and embeddings, in place of a fixed basis. A future direction is to constrain the learned basis, for example, with orthonormality~\citep{ramsay2005functional}.

Other neural forecast architectures scratched the surface of the \ours forecast strategy, examples include \DAM~\citep{darlow2024dam} and \NBEATS~\citep{oreshkin2020nbeats, olivares2021nbeatsx} that uses harmonics and polynomial trends, \NHITS~\citep{challu2023nhits} that use wavelets. In contrast to these approaches, we do not pre-conditioning the function family and rather learn it via embeddings, but more importantly we allow for variable length horizon in contrast to these architectures that have \MS fixed outputs.

Neural DEs~\citep{chen2018neural, rubanova2019latent, kidger2020neural} learn vector fields that are integrated to obtain continuous-horizon trajectories. In contrast, \ours\ retains a discrete-time encoder and uses a horizon-separable basis for direct, parallel evaluation without numerical integration. Inspired by ODEs and integrated time-series models (e.g., ARIMA), a promising direction is to apply \ours\ to first differences and integrate predictions to recover forecast levels, potentially improving robustness to non-stationary trends.

Our simulations decompose only the point-forecast errors. A probabilistic extension can use the decomposition of \citet{hersbach2000decomposition}, to study the uncertainty, calibration, and reliability tradeoffs of different forecast strategies.

\citet{taieb2015bias} proposed discrete hybrid strategies, such as DirRec, that combine the low variance of recursive forecasts with the low bias of direct forecasts. A hybrid of discrete strategies and \ours\ is a further direction for very long horizons.

Current foundation forecasting models use a fixed output grid or a recursive rollout~\citep{garza2023timegpt, aws2024chronos, google2024timesfm, salesforce2023moirai, hoo2024tabpfn} and truncate a long output to get a shorter horizon. \ours\ gives only the necessary horizons, so its integration with foundation models is a promising direction.

 \clearpage
\section{Conclusion} \label{sec:08_conclusion}

We introduced the \ourscomplete (\ours) forecast strategy, and conducted a comprehensive theoretical investigation of the properties of \ours, showing that it provides universal approximation guaranties for forecasting operators. Furthermore, we proved that, under reasonable assumptions, the estimation variance of \ours improves on that of discrete strategies because \ours makes a more efficient use of the information from the horizon. 

We validated these theoretical findings through simulation studies and bias-variance decompositions, demonstrating that \ours achieves a bias profile comparable to \DIR and \MS with a lower estimation variance. Beyond fixed grid training, \ours demonstrates strong accuracy in both interpolation and extrapolation regimes, forecasting horizons unseen during training. 

\ours achieves competitive accuracy. It has a lower relAvgRMSE than \MS\ in 40 of the 64 encoder and frequency pairs of the M-series benchmark. Finally, we highlight the substantial computational efficiency of \ours, particularly its ability to execute targeted horizon queries in constant time.


\bibliographystyle{elsarticle-harv}
\bibliography{refs}

\appendix \newpage

\section{Universal Operator Approximation Theorem}
\label{appendix:universal_approximation}

In this appendix, we prove the \emph{universal approximation theorem} introduced in Section~\ref{sec:03_theory}. We use the Stone-Weierstrass theorem to prove the density of finite sum approximation functions, and complement with the universal approximation theorem applied to its components.


\vspace{.5cm}

\begin{lemma}\label{lem:separable_density} Let $\mathcal{Y} \subset \mathbb{R}^{t}$ be a compact domain and $[0,H] \subset \mathbb{R}^+$ a compact horizon interval. Let $f : \mathcal{Y} \times [0,H] \to \mathbb{R}$ given by $(\mathbf{y}_{:t}, h) \mapsto f(h \mid \mathbf{y}_{:t})$ be a continuous forecast function. Then, for any $\epsilon > 0$, there exists a finite integer $K \in \mathbb{N}$ and continuous functions $\btheta_k^\star : \mathcal{Y} \to \mathbb{R}$ and $\bphi_k^\star : [0,H] \to \mathbb{R}$ for $k = 1, \dots, K$ such that:

$$\sup_{\mathbf{y}_{:t} \in \mathcal{Y}, \, h \in [0,H]} \left\vert{} f(h \mid \mathbf{y}_{:t}) - \sum_{k=1}^K \btheta_k^\star(\mathbf{y}_{:t}) \bphi_k^\star(h) \right\vert{} < \epsilon$$

\end{lemma}
\begin{proof} The Cartesian product of two compact sets $\mathcal{Y} \times [0,H]$ is compact in $\mathbb{R}^{t+1}$. Consider the set of finite separable functions:

$$\mathcal{A} = \left\{ \sum_{k=1}^K \btheta_k(\mathbf{y}_{:t}) \bphi_k(h) \;:\; K < \infty, \; \btheta_k \in \mathcal{C}(\mathcal{Y}), \; \bphi_k \in \mathcal{C}([0,H]) \right\}$$

We verify that $\mathcal{A}$ forms a subalgebra of $\mathcal{C}(\mathcal{Y} \times [0,H])$ that satisfies the conditions of the Stone-Weierstrass Theorem.
\begin{enumerate}
    \item Contains Constants: Taking $\btheta(\mathbf{y}_{:t}) = 1$ and $\bphi(h) = c$ shows $c \in \mathcal{A}$.
    \item Closed under Addition and Multiplication: For any $g_1, g_2 \in \mathcal{A}$, $g_1 + g_2 \in \mathcal{A}$ and $g_1 g_2 \in \mathcal{A}$ by expansion of finite sums.
    \item Point Separation: For $(\mathbf{y}, h) \neq (\mathbf{y}', h')$, if $\mathbf{y} \neq \mathbf{y}'$, choose $\btheta(\mathbf{y}) = y_i$ (where coordinate $y_i \neq y'_i$) and $\bphi(h) = 1$. If $h \neq h'$, choose $\btheta(\mathbf{y}) = 1$ and $\bphi(h) = h$.
\end{enumerate}

By the Stone-Weierstrass Theorem, $\mathcal{A}$ is dense in $\mathcal{C}(\mathcal{Y} \times [0,H])$ equipped with the uniform norm. Therefore, for any $f \in \mathcal{C}(\mathcal{Y} \times [0,H])$ and any $\epsilon > 0$, there exists some element in $\mathcal{A}$ within the distance $\epsilon$ of $f$, establishing the existence of finite $K$ and continuous functions $\{\btheta_k^\star\}_{k=1}^K, \{\bphi_k^\star\}_{k=1}^K$.

\end{proof}

\clearpage
\begin{lemma}\label{lem:basis_uat}
Let $\{\bphi_k^{\star}\}_{k=1}^{K}$ be a finite set of continuous functions
\[
\bphi_k^{\star}:[0,H]\rightarrow\mathbb{R},
\]
where $[0,H]\subset\mathbb{R}^{+}$ is compact. Then, for all $\epsilon>0$, there exist neural networks
\[
\{\bphi_k\}_{k=1}^{K}
\]
such that
\begin{align}
    \sup_{h\in[0,H]}
    \left|
    \bphi_k^{\star}(h)
    -
    \bphi_k(h)
    \right|
    <\epsilon,
    \qquad
    k=1,\dots,K.
\end{align}

\end{lemma}
\begin{proof}
This lemma is a direct consequence of the neural universal approximation theorem~\citep{hornik1991approximation}. Since every $\bphi_k^{\star}$ is continuous in the compact domain $[0,H]$, a non-polynomial activated feedforward neural network can approximate $\bphi_k^{\star}$ uniformly to arbitrary precision. As the number of functions $K$ is finite, the approximation can be constructed for every $k\in\{1,\dots,K\}$ using the same target precision $\epsilon$.
\end{proof}

\begin{lemma}\label{lem:coef_uat}
Let $\{\btheta_k^{\star}\}_{k=1}^{K}$ be a finite set of continuous coefficient mappings
\[
\btheta_k^{\star}:\mathcal{Y}\rightarrow\mathbb{R},
\]
where $\mathcal{Y}\subset\mathbb{R}^{t}$ is compact. Then, for all $\epsilon>0$, there exist neural networks
\[
\{\btheta_k\}_{k=1}^{K}
\]
such that
\begin{align}
    \sup_{\mathbf{y}_{:t}\in\mathcal{Y}}
    \left|
    \btheta_k^{\star}(\mathbf{y}_{:t})
    -
    \btheta_k(\mathbf{y}_{:t})
    \right|
    <\epsilon,
    \qquad
    k=1,\dots,K.
\end{align}

\end{lemma}
\begin{proof}
As in Lemma~\ref{lem:basis_uat}, this follows from the neural universal approximation theorem. Since every $\btheta_k^{\star}$ is continuous in the compact domain $\mathcal{Y}$, there exists a neural network $\btheta_k$ that approximates it uniformly with arbitrary precision. Since $K$ is finite, the approximation can be constructed simultaneously for all coefficient mappings.
\end{proof}

\clearpage
\noindent \textbf{Theorem 1.}
Let $\mathcal{Y}\subset\mathbb{R}^{t}$ be a compact domain of historical observations $\mathbf{y}_{:t}$ and let $[0,H]\subset\mathbb{R}^{+}$ be a compact forecast horizon interval. Let a continuous forecast operator be

$$
F:\mathcal{Y}\rightarrow\mathcal{C}([0,H]),
\qquad
F(\mathbf{y}_{:t})=f(\cdot\mid\mathbf{y}_{:t}),
$$

where $\mathcal{C}([0,H])$ is equipped with the uniform norm. Then, for all $\epsilon>0$, there exists a finite $K\in\mathbb{N}$, a set of continuous neural network basis functions $\{\bphi_{k}(h)\}_{k=1}^{K}$, and neural coefficient mappings $\{\btheta_{k}(\mathbf{y}_{:t})\}_{k=1}^{K}$, collected in the parameter vector $\mathbf{w}^{\star}$, such that the neural forecaster $g(h\mid\mathbf{y}_{:t},\mathbf{w}^{\star})=\sum_{k=1}^{K}\btheta_{k}(\mathbf{y}_{:t})\,\bphi_{k}(h)$ approximates $f$ uniformly over the horizon,
\begin{equation}
\sup_{h\in[0,H]} \left| f\left(h \mid \mathbf{y}_{:t}\right) - g\left(h \mid \mathbf{y}_{:t}, \mathbf{w}^{\star}\right) \right| < \epsilon, \quad \forall \mathbf{y}_{:t} \in \mathcal{Y}.
\end{equation}
Because $[0,H]$ has finite length, this uniform bound implies the squared-error bound
\begin{equation}
\int_0^H \left( f\left(h \mid \mathbf{y}_{:t}\right) - g\left(h \mid \mathbf{y}_{:t}, \mathbf{w}^{\star}\right) \right)^2 dh < H\,\epsilon^2, \quad \forall \mathbf{y}_{:t} \in \mathcal{Y},
\end{equation}
which is the form that controls the squared approximation bias $B_h$ of Lemma~\ref{lem:taieb_mse}.

\begin{proof}
Since
\[
F:\mathcal{Y}\rightarrow\mathcal{C}([0,H])
\]
is continuous under the uniform norm, the associated joint forecast function
\[
f:\mathcal{Y}\times[0,H]\rightarrow\mathbb{R},
\qquad
(\mathbf{y}_{:t},h)\mapsto f(h\mid\mathbf{y}_{:t}),
\]
is continuous.

Using Lemma~\ref{lem:separable_density}, for all $\epsilon_1>0$, there exist finite $K$ and continuous functions
$\btheta_k^{\star}$ and $\bphi_k^{\star}$ such that
\begin{align}
    \sup_{\mathbf{y}_{:t}\in\mathcal{Y}}
    \sup_{h\in[0,H]}
    \left|
    f(h\mid\mathbf{y}_{:t})
    -
    \sum_{k=1}^{K}
    \btheta_k^{\star}(\mathbf{y}_{:t})
    \bphi_k^{\star}(h)
    \right|
    <\epsilon_1.
\end{align}

Let
\[
g_K^{\star}(h\mid\mathbf{y}_{:t})
=
\sum_{k=1}^{K}
\btheta_k^{\star}(\mathbf{y}_{:t})
\bphi_k^{\star}(h)
\]
denote this finite separable approximation.

We now approximate the two ingredients of $g_K^{\star}$.

Using Lemma~\ref{lem:basis_uat}, for all $\epsilon_2>0$, there exist basis functions
$\bphi_k$ such that
\begin{align}
    \sup_{h\in[0,H]}
    |\bphi_k^{\star}(h)-\bphi_k(h)|
    <\epsilon_2.
\end{align}

Using Lemma~\ref{lem:coef_uat}, for all $\epsilon_3>0$, there exist coefficients
$\btheta_k$ such that
\begin{align}
    \sup_{\mathbf{y}_{:t}\in\mathcal{Y}}
    |\btheta_k^{\star}(\mathbf{y}_{:t})
    -
    \btheta_k(\mathbf{y}_{:t})|
    <\epsilon_3.
\end{align}

Define the neural forecast approximation as
\[
g(h\mid\mathbf{y}_{:t},\mathbf{w}^{\star})
=
\sum_{k=1}^{K}
\btheta_k(\mathbf{y}_{:t})
\bphi_k(h),
\]
where the single finite parameter vector $\mathbf{w}^{\star} = \big(\mathbf{w}^{\star}_{\btheta,1},\dots,\mathbf{w}^{\star}_{\btheta,K},\, \mathbf{w}^{\star}_{\bphi,1},\dots,\mathbf{w}^{\star}_{\bphi,K}\big)$ concatenates the weights of the $K$ coefficient networks $\{\btheta_k\}_{k=1}^{K}$ (each with weights $\mathbf{w}^{\star}_{\btheta,k}$) and the $K$ basis networks $\{\bphi_k\}_{k=1}^{K}$ (each with weights $\mathbf{w}^{\star}_{\bphi,k}$). Since $K$ is finite and each network has finitely many weights, $\mathbf{w}^{\star}$ is finite-dimensional.

To conclude, we combine the finite separable approximation, the neural basis approximation, and the neural coefficient approximation.

Using the triangular inequality,
\begin{align}
\begin{split}
&\sup_{h\in[0,H]}
\left|
f(h\mid\mathbf{y}_{:t})
-
g(h\mid\mathbf{y}_{:t},\mathbf{w}^{\star})
\right|
\\
&\leq
\sup_{h\in[0,H]}
\left|
f(h\mid\mathbf{y}_{:t})
-
\sum_{k=1}^{K}
\btheta_k^{\star}(\mathbf{y}_{:t})
\bphi_k^{\star}(h)
\right|
\\
&\quad+
\sup_{h\in[0,H]}
\left|
\sum_{k=1}^{K}
\btheta_k^{\star}(\mathbf{y}_{:t})
\bphi_k^{\star}(h)
-
\sum_{k=1}^{K}
\btheta_k^{\star}(\mathbf{y}_{:t})
\bphi_k(h)
\right|
\\
&\quad+
\sup_{h\in[0,H]}
\left|
\sum_{k=1}^{K}
\btheta_k^{\star}(\mathbf{y}_{:t})
\bphi_k(h)
-
\sum_{k=1}^{K}
\btheta_k(\mathbf{y}_{:t})
\bphi_k(h)
\right|.
\end{split}
\end{align}

Bounding each supremum through the triangle inequality over the sum,
\begin{align}
\begin{split}
&\sup_{h\in[0,H]}
\left|
f(h\mid\mathbf{y}_{:t})
-
g(h\mid\mathbf{y}_{:t},\mathbf{w}^{\star})
\right|
\\
&\leq
\epsilon_1
+
\sum_{k=1}^{K}
|\btheta_k^{\star}(\mathbf{y}_{:t})|
\sup_{h\in[0,H]}
|\bphi_k^{\star}(h)-\bphi_k(h)|
\\
&\quad+
\sum_{k=1}^{K}
|\btheta_k^{\star}(\mathbf{y}_{:t})
-
\btheta_k(\mathbf{y}_{:t})|
\sup_{h\in[0,H]}
|\bphi_k(h)|.
\end{split}
\end{align}

Using continuity in compact domains, the finite collections
$\{\btheta_k^{\star}\}$ and $\{\bphi_k\}$ are uniformly bounded. Let
\[
M_{\theta}
=
\max_{1\leq k\leq K}
\sup_{\mathbf{y}_{:t}\in\mathcal{Y}}
|\btheta_k^{\star}(\mathbf{y}_{:t})|,
\quad \text{and} \quad 
M_{\phi}
=
\max_{1\leq k\leq K}
\sup_{h\in[0,H]}
|\bphi_k(h)|.
\]

Then
\begin{align}
\begin{split}
&\sup_{h\in[0,H]}
\left|
f(h\mid\mathbf{y}_{:t})
-
g(h\mid\mathbf{y}_{:t},\mathbf{w}^{\star})
\right| <
\epsilon_1
+
K M_{\theta}\epsilon_2
+
K M_{\phi}\epsilon_3.
\end{split}
\end{align}

Finally, the three approximation accuracies can be chosen
\[
\epsilon_1
=
\frac{\epsilon}{3},
\qquad
\epsilon_2
=
\frac{\epsilon}
{3K M_{\theta}},
\qquad
\epsilon_3
=
\frac{\epsilon}
{3K M_{\phi}},
\]
which gives the uniform bound
\begin{align}
    \sup_{h\in[0,H]}
    \left|
    f(h\mid\mathbf{y}_{:t})
    -
    \sum_{k=1}^{K}
    \btheta_k(\mathbf{y}_{:t})
    \bphi_k(h)
    \right|
    <\epsilon,
    \qquad
    \forall\mathbf{y}_{:t}\in\mathcal{Y}.
\end{align}

Since $[0,H]$ has finite length $H$, the uniform bound controls both the $L_1$ error and the squared error,
\begin{align}
\begin{split}
    &\int_0^H
    \left|
    f(h\mid\mathbf{y}_{:t})
    -
    g(h\mid\mathbf{y}_{:t},\mathbf{w}^{\star})
    \right|dh
    < H\epsilon,
    \\
    &\qquad\qquad
    \int_0^H
    \left(
    f(h\mid\mathbf{y}_{:t})
    -
    g(h\mid\mathbf{y}_{:t},\mathbf{w}^{\star})
    \right)^2 dh
    < H\epsilon^2,
\end{split}
\end{align}
for all $\mathbf{y}_{:t}\in\mathcal{Y}$. The squared-error bound is the form that controls the squared approximation bias $B_h$ of Lemma~\ref{lem:taieb_mse}.

Therefore, the continuous operator can be arbitrarily approximated by a finite neural basis expansion whose basis functions depend on the forecast horizon and whose coefficients depend on the historical observations.
\end{proof} \clearpage
\section{Continuous Horizon Variance Dominance}
\label{appendix:ch_decomposition}

In this appendix, we prove the \emph{variance dominance theorem} introduced in Section~\ref{sec:03_theory}, which bounds the estimation variance $V_h$ of Lemma~\ref{lem:taieb_mse} across forecast strategies. The argument proceeds in two stages. First, we characterize how the horizon-selector measure $\rho$ shapes the covariance of the fitted parameters: a first-order expansion of the empirical-risk minimizer yields a sandwich covariance $\operatorname{Cov}(\hat{\mathbf w}) \approx \tfrac{1}{T} A^{-1}\Omega(\rho) A^{-1}$, whose variability matrix $\Omega(\rho)$ aggregates the cross-horizon residual covariances selected by $\rho$, and the delta method propagates the parameter covariance to the horizon-$h$ forecast covariance. Second, we turn the forecast covariance into a scalar ordering: under a separable score covariance with a correlation kernel $K$, the variability collapses to a single \emph{effective horizon information} $\mathcal I_{\mathrm{eff}}(\rho)$, the number of independent horizons the forecast strategy's objective supplies during estimation. Direct forecasting draws on one horizon, multi-step on a discrete grid, and \ourscomplete on the whole interval, so their effective information obeys $\mathcal I_{\mathrm{eff}}^{\mathrm{DIR}} < \mathcal I_{\mathrm{eff}}^{\mathrm{MS}} < \mathcal I_{\mathrm{eff}}^{\mathrm{CH}}$ at any finite grid resolution. Because variance scales inversely with $\mathcal I_{\mathrm{eff}}$ at matched shared-parameter dimension, this chain gives the variance ordering $V_{\mathrm{CH}}(h) < V_{\mathrm{MS}}(h) < V_{\mathrm{DIR}}(h)$.

\subsection{Parameters Covariance Characterization} \label{appendix:ch_variance}

The variance term $V_h$ of Lemma~\ref{lem:taieb_mse} is governed by the covariance of the fitted parameters, which the following lemma characterizes for \REC, \DIR, \MS, and \ours strategies at once. Each strategy is identified by the horizon-selector measure $\rho$ of Eq.~\eqref{eq:erm}: a point mass $\delta_h$ for the direct strategy, the empirical measure $\tfrac{1}{H}\sum_j \delta_{h_j}$ for multi-step, and the uniform measure $\tfrac{1}{H}\,dh$ for the \ourscomplete strategy.

\begin{lemma}[Parameters' covariance under a horizon selector]
\label{lem:cov}
Let $\hat{\mathbf w}$ minimize the $\rho$-averaged empirical risk of Eq.~\eqref{eq:erm} over the $T$ tuples of $\mathcal{D}$. Define the per-horizon score at the population optimum $\mathbf w^\star$ and its $\rho$-aggregate,
\[
s(h) = \nabla_{\mathbf w} L\big(y_{t+h},\, g(h\mid\mathbf{y}_{:t},\mathbf w^\star)\big),
\qquad
S = \int_0^H s(h)\,d\rho(h).
\]
The estimator $\hat{\mathbf w}$ solves the first-order condition,  linearizing around $\mathbf w^\star$,
\[
\frac1T\sum_{t=1}^{T} S_t(\hat{\mathbf w}) = 0,
\qquad
0 \approx \frac1T\sum_{t=1}^{T} S_t(\mathbf w^\star) + A\,(\hat{\mathbf w}-\mathbf w^\star),
\]
where $S_t = \nabla_{\mathbf w} R_t$ is the score of tuple $t$, the gradient of its per-tuple risk $R_t(\mathbf w) = \int_0^H L\big(y_{t+h}, g(h\mid\mathbf{y}_{:t},\mathbf w)\big)\,d\rho(h)$, and $A = \mathbb E[\nabla_{\mathbf w}^2 R_t(\mathbf w^\star)]$ is the expected Hessian of the risk at the optimum. Assuming (a) $\hat{\mathbf w}$ is consistent for $\mathbf w^\star$, (b) $A$ is invertible, and (c) the expansion remainder is asymptotically negligible, we solve the linear system
\[
\hat{\mathbf w} - \mathbf w^\star \approx -A^{-1}\,\tfrac1T\sum_t S_t(\mathbf w^\star).
\]
Its right-hand side is an average of $T$ mean-zero scores with covariance $\tfrac1T\Omega(\rho)$, $\Omega(\rho) = \operatorname{Cov}(S)$, by bilinearity of the covariance operator. Propagating through $A^{-1}$~\citep{huber1967behavior, vandervaart1998asymptotic}, the fitted parameters therefore have covariance
\begin{equation}
\operatorname{Cov}(\hat{\mathbf w})
\approx
\frac{1}{T}\, A^{-1}\,\Omega(\rho)\,A^{-1},
\label{eq:cov_sandwich}
\end{equation}
with variability matrix $\Omega(\rho)$ given by
\begin{equation}
\Omega(\rho) = \operatorname{Cov}(S)
= \int_0^H\!\int_0^H \operatorname{Cov}\big(s(h),s(h')\big)\,d\rho(h)\,d\rho(h').
\label{eq:cov_AOmega}
\end{equation}
\end{lemma}

Substituting each selector $\rho$ into Eq.~\eqref{eq:cov_AOmega} yields its variability matrix.

\begin{proposition}[Per-strategy variability]
\label{prop:per_strategy}
With the horizon selectors of Eq.~\eqref{eq:erm}, the variability matrix $\Omega(\rho)$ takes the forms
\begin{align}
\Omega(\rho_{\mathrm{DIR}}) &= \operatorname{Var}\!\big(s(h)\big), \label{eq:omega_dir}\\
\Omega(\rho_{\mathrm{MS}}) &= \frac{1}{H^2}\sum_{h=1}^{H}\sum_{h'=1}^{H} \operatorname{Cov}\!\big(s(h),s(h')\big), \label{eq:omega_ms}\\
\Omega(\rho_{\mathrm{CH}}) &= \frac{1}{H^2}\int_0^H\!\int_0^H \operatorname{Cov}\!\big(s(h),s(h')\big)\,dh\,dh'. \label{eq:omega_ch}
\end{align}
The direct strategy estimates an independent parameter block $\mathbf w_h$ for each horizon, so its selector $\rho=\delta_h$ retains only the diagonal term and no cross-horizon covariance enters; multi-step and the \ourscomplete strategy pool a single $\mathbf w$ across horizons, so their $\Omega(\rho)$ accumulate the off-diagonal score covariances over the discrete grid and the continuum, respectively.
\end{proposition}

The three forms include progressively more off-diagonal score covariance: none for direct, a discrete double sum over the $H$ grid points for multi-step, and the full double integral over the continuum for \ourscomplete. The estimator covariance of Lemma~\ref{lem:cov}, with the per-strategy $\Omega(\rho)$ of Proposition~\ref{prop:per_strategy}, propagates to the forecast through the delta method, giving the horizon-$h$ estimation variance $V_h$ of Lemma~\ref{lem:taieb_mse}.

\begin{corollary}[Forecast variance]
\label{cor:delta_variance}
Let $\psi = \nabla_{\mathbf w} g(h\mid\mathbf{y}_{:t},\mathbf w^\star)$ be the forecast gradient at the optimum. Linearization of the forecast around $\mathbf w^\star$,
\[
g(h\mid\mathbf{y}_{:t},\hat{\mathbf w})
\approx
g(h\mid\mathbf{y}_{:t},\mathbf w^\star)
+ \psi^\top (\hat{\mathbf w}-\mathbf w^\star),
\]
so that, substituting the covariance of Eq.~\eqref{eq:cov_sandwich},
\begin{equation}
V_h(\mathbf{y}_{:t})
\approx
\psi^\top \operatorname{Cov}(\hat{\mathbf w})\, \psi
=
\frac{1}{T}\,
\psi^\top A^{-1}\Omega(\rho) A^{-1} \psi.
\label{eq:delta_variance}
\end{equation}
\end{corollary}

The forecast variance thus inherits the sandwich structure, quadratic in the forecast gradient $\psi$ and scaled by the parameter covariance that the selector $\rho$ shapes through $A$ and $\Omega(\rho)$.

\subsection{Effective horizon information}

To turn the variability matrix into a scalar measure of how much independent information the horizons provide, we impose a separable structure on the score covariance.

\begin{assumption}[Separable score covariance]
\label{ass:separable}
The horizon-score process has covariance
\begin{equation}
\operatorname{Cov}\big(s(h), s(h')\big) = G\, K(h,h'),
\label{eq:separable}
\end{equation}
where $G \succeq 0$ is a fixed parameter-space matrix and $K$ is a correlation kernel: continuous, positive semidefinite, with $K(h,h)=1$ and $0 \le K(h,h') \le 1$, decaying in $|h-h'|$ with finite correlation length.
\end{assumption}

The nonnegative, decaying correlation captures that nearby forecast horizons provide overlapping rather than independent information. This is the score-space analogue of the effective-sample-size construction in spatial statistics~\citep{griffith2005effective, vallejos2014effective}, and of its functional extension~\citep{acosta2026functional}, where correlated observations carry less information than their nominal count.

\begin{lemma}[Effective horizon information]
\label{thm:eff_horizon}
Under Assumption~\ref{ass:separable}, define the effective horizon information of a strategy with selector $\rho$ as
\begin{equation}
\mathcal I_{\mathrm{eff}}(\rho)
=
\left[ \int_0^H\!\int_0^H K(h,h')\, d\rho(h)\, d\rho(h') \right]^{-1}.
\label{eq:ieff}
\end{equation}
Then the variability matrix factors as $\Omega(\rho) = G\,/\,\mathcal I_{\mathrm{eff}}(\rho)$, and the shared-parameter covariance of Lemma~\ref{lem:cov} becomes
\begin{equation}
\operatorname{Cov}(\hat{\mathbf w}_\rho)
\approx
\frac{1}{T\,\mathcal I_{\mathrm{eff}}(\rho)}\, A_\rho^{-1} G A_\rho^{-1}.
\label{eq:cov_ieff}
\end{equation}
\end{lemma}

\begin{corollary}[Effective information per strategy]
\label{cor:ieff_strategies}
Substituting the selectors of Eq.~\eqref{eq:erm} into Eq.~\eqref{eq:ieff},
\begin{equation}
\mathcal I_{\mathrm{eff}}^{\mathrm{DIR}} = 1,
\qquad
\mathcal I_{\mathrm{eff}}^{\mathrm{MS}} = \frac{H^2}{\sum_{h}\sum_{h'} K(h,h')},
\qquad
\mathcal I_{\mathrm{eff}}^{\mathrm{CH}} = \frac{H^2}{\int_0^H\!\int_0^H K(h,h')\,dh\,dh'}.
\label{eq:ieff_strategies}
\end{equation}

These satisfy two limits. When the horizons are fully correlated ($K\equiv 1$), every strategy collapses to $\mathcal I_{\mathrm{eff}}=1$: all horizons carry the information of one. When the multi-step horizons are independent ($K(h,h')=\mathbbm{1}\{h=h'\}$), $\mathcal I_{\mathrm{eff}}^{\mathrm{MS}}=H$, the nominal count underlying the residuals-per-parameter argument of~\citet{taieb2015bias}; for positively correlated horizons $1\le \mathcal I_{\mathrm{eff}}^{\mathrm{MS}}\le H$. The independent case is thus a special case of multi-step, not an assumption imposed on \ourscomplete.
\end{corollary}

For a stationary kernel $K(h,h')=K(u)$ with lag $u=|h-h'|$, when $H$ is large relative to the correlation length the \ourscomplete effective information admits the closed form
\begin{equation}
\mathcal I_{\mathrm{eff}}^{\mathrm{CH}} \approx \frac{H}{2\,\ell_{\mathrm{int}}},
\qquad
\ell_{\mathrm{int}} = \int_0^\infty K(u)\,du.
\label{eq:ieff_ch_limit}
\end{equation}
Information grows with the interval length divided by the correlation length; note that a denser horizon sampling does not drive the variance to zero while $K$ remains positively correlated.

\begin{remark}[Residuals per parameter]
\label{rem:rpp}
Under the additional isotropy assumption that the horizon-averaged sensitivity factor $\psi^\top A_\rho^{-1} G A_\rho^{-1} \psi$ scales with the number of shared parameters $P_\rho$, the forecast variance takes the residuals-per-parameter form
\begin{equation}
V_\rho(h) \;\propto\; \frac{P_\rho}{T\,\mathcal I_{\mathrm{eff}}(\rho)},
\qquad
\mathrm{RPP}(\rho) \;\propto\; \frac{T\,\mathcal I_{\mathrm{eff}}(\rho)}{P_\rho}.
\label{eq:rpp}
\end{equation}
\end{remark}

The sharpest comparison is with multi-step, since both strategies pool a single $\mathbf w$ and differ only in how the horizon objective is sampled: a discrete grid of $H$ points for multi-step, the whole interval for \ourscomplete. Their effective information is governed by the same kernel $K$, so $\mathcal I_{\mathrm{eff}}^{\mathrm{MS}}$ is the Riemann sum that $\mathcal I_{\mathrm{eff}}^{\mathrm{CH}}$ integrates, and refining the grid past the correlation length adds only correlated scores, so $\mathcal I_{\mathrm{eff}}^{\mathrm{MS}}$ saturates at the continuum ceiling rather than growing with the nominal count. This is the continuous form of the classical bias--variance trade-off: the pooling that lowers the variance is bounded by the horizon correlation structure, and \ourscomplete realizes that bound without committing to a fixed output grid.

\noindent \textbf{Theorem 2.} (Continuous Horizon Variance Dominance). 

Suppose, in addition to Assumption~\ref{ass:separable}, that the stationary correlation kernel $K(u)$ is nonincreasing and strictly convex in $|u|$, and that the multi-step horizons form a regular grid on $[0,H]$. Then the continuous horizon, multi-step, and direct forecast strategies' effective information satisfy
\begin{equation}
1 = \mathcal I_{\mathrm{eff}}^{\mathrm{DIR}} <\mathcal I_{\mathrm{eff}}^{\mathrm{MS}} \;<\; \mathcal I_{\mathrm{eff}}^{\mathrm{CH}},
\label{eq:ch_dominates}
\end{equation}
with equality between multi-step and continuous horizon only in the fine-grid limit. Consequently, for a matched capacity of shared parameters $P$ (the shared trunk parameters, held fixed between multi-step and \ourscomplete so that only the horizon sampling differs), the forecast estimation variance satisfies
\begin{equation}
V_{\mathrm{CH}}(h) < V_{\mathrm{MS}}(h) < V_{\mathrm{DIR}}(h), \quad \forall h \in [0, H],
\label{eq:ch_variance_order}
\end{equation}
so the continuous-horizon estimator attains strictly more variance reduction than any regular multi-step grid at any finite resolution.

\noindent\emph{Proof sketch.} The direct strategy carries $\mathcal I_{\mathrm{eff}}^{\mathrm{DIR}}=1$ by Corollary~\ref{cor:ieff_strategies}, while any positively correlated grid of more than one horizon gives $\mathcal I_{\mathrm{eff}}^{\mathrm{MS}}>1$, establishing the strict left inequality. For the right inequality, by Eq.~\eqref{eq:ieff} $\mathcal I_{\mathrm{eff}}^{\mathrm{MS}} < \mathcal I_{\mathrm{eff}}^{\mathrm{CH}}$ is equivalent to the grid average of $K$ exceeding its continuum average. For a strictly convex, nonincreasing $K$ on a regular grid, the rectangle rule strictly underestimates the integral of $K$ over each cell, so the grid double-average strictly dominates the double integral; inverting reverses the inequality. As the grid spacing vanishes, the rectangle rule converges to the integral, so equality holds only in the fine-grid limit. The variance ordering of Eq.~\eqref{eq:ch_variance_order} then follows from Eq.~\eqref{eq:cov_ieff}: at the matched shared-parameter capacity $P$, the forecast variance scales as $1/\mathcal I_{\mathrm{eff}}(\rho)$, so the information ordering inverts into the stated variance ordering. \hfill$\square$

 \clearpage
\section{MSE Decomposition Experiment Details}
\label{appendix:dpgs_details}

\subsection{Bias and Variance Analysis by Simulation}

The bias and variance study of Section~\ref{sec:04_theory_validation} uses the three autoregressive processes of \citet{taieb2015bias}. Each process has i.i.d.\ Gaussian innovations $\varepsilon_t$.

\textbf{Smooth Transition Autoregression (STAR).} This process has two regimes. A logistic function of $y_{t-1}$ moves it smoothly between them~\citep{terasvirta2010modelling}:
    \begin{equation}
    y_t = 0.3\,y_{t-1} + 0.6\,y_{t-2} + \big(0.1 - 0.9\,y_{t-1} + 0.8\,y_{t-2}\big)\big[1 + e^{-10\,y_{t-1}}\big]^{-1} + \varepsilon_t,
    \label{eq:dgp_star}
    \end{equation}
with $\varepsilon_t \sim \mathcal{N}(0, \sigma^2)$ and $\sigma = 0.01$. Thus, the process is almost deterministic.

\textbf{Nonlinear Autoregression (NLAR).} This process is a neural network with two hidden units, fit to the annual sunspot series~\citep{terasvirta2010modelling}. It has a linear part and two logistic gates:
    \begin{equation}
    \begin{aligned}
    y_t &= -0.17 + 0.85\,y_{t-1} + 0.14\,y_{t-2} - 0.31\,y_{t-3} + 0.08\,y_{t-7} \\
    &\quad + 12.80\, G_1(y_{t-1}) + 2.44\, G_2(y_{t-1}) + \varepsilon_t,
    \end{aligned}
    \label{eq:dgp_nlar}
    \end{equation}
    \begin{align}
    G_1(y_{t-1}) &= \big(1 + \exp\{-0.46(0.29\,y_{t-1} - 0.87\,y_{t-2} + 0.40\,y_{t-7} - 6.68)\}\big)^{-1}, \\
    G_2(y_{t-1}) &= \big(1 + \exp\{-1.17\times10^{3}(0.83\,y_{t-1} - 0.53\,y_{t-2} - 0.18\,y_{t-7} + 0.38)\}\big)^{-1},
    \end{align}
with $\varepsilon_t \sim \mathcal{N}(0, 1)$.

\textbf{Linear Autoregression (LAR).} This process is an AR(6) fit to the annual sunspot series~\citep{box2015time}:
    \begin{equation}
    y_t = 1.32\,y_{t-1} - 0.52\,y_{t-2} - 0.16\,y_{t-3} + 0.18\,y_{t-4} - 0.26\,y_{t-5} + 0.19\,y_{t-6} + \varepsilon_t,
    \label{eq:dgp_lar}
    \end{equation}
with $\varepsilon_t \sim \mathcal{N}(0, 1)$. The process is linear, so $\sigma$ only scales the series. It shows the cost of a flexible decoder when the true process is linear.

\textbf{Training.} Each training sample has $N=8$ short series. For each process, we discard the first 300 simulated values. We train each strategy for 2,500 to 3,000 steps on each sample. We use $L=80$ independent samples for the bias and variance estimates, and we report the horizons up to $H=20$. The series length changes as in \citet{taieb2015bias}.

\clearpage
\subsection{NLAR MSE Decomposition} \label{appendix:nlar_decomposition}

\vspace{4cm}
\begin{figure}[ht!]
    \centering
    \includegraphics[width=\linewidth]{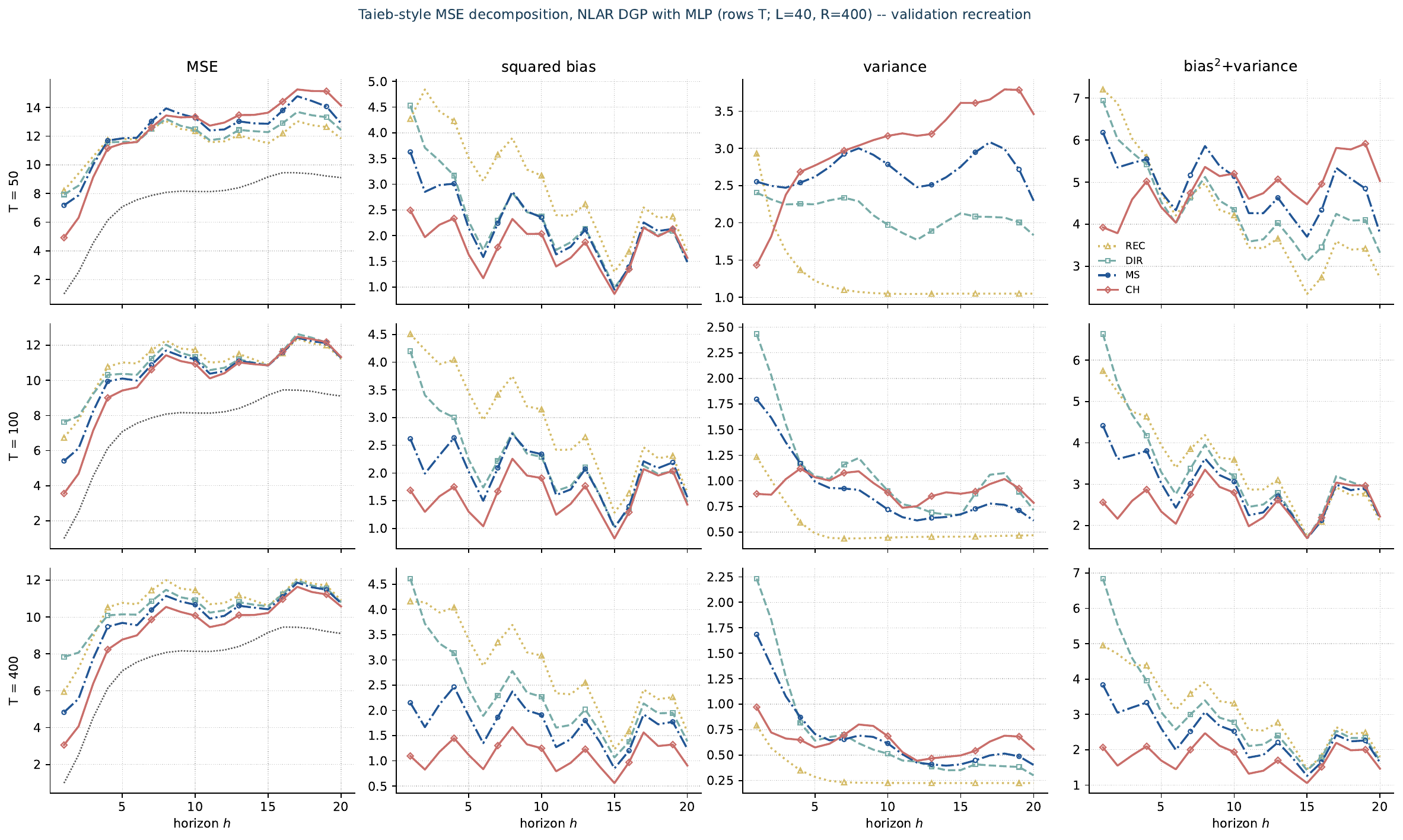}
    \caption{MSE decomposition on the NLAR process~\citep{taieb2015bias} across training-series lengths $T$ and horizons $h$. Direct and multi-step strategies reproduce the classical bias--variance trade-off, while \ourscomplete (\ours) attains the lowest total error, matching multi-step at short lengths and pulling ahead as $T$ increases.}
    \label{fig:taieb_nlar}
\end{figure}

\clearpage
\subsection{STAR MSE Decomposition} \label{appendix:star_decomposition}

\vspace{4cm}
\begin{figure}[ht!]
    \centering
    \includegraphics[width=\linewidth]{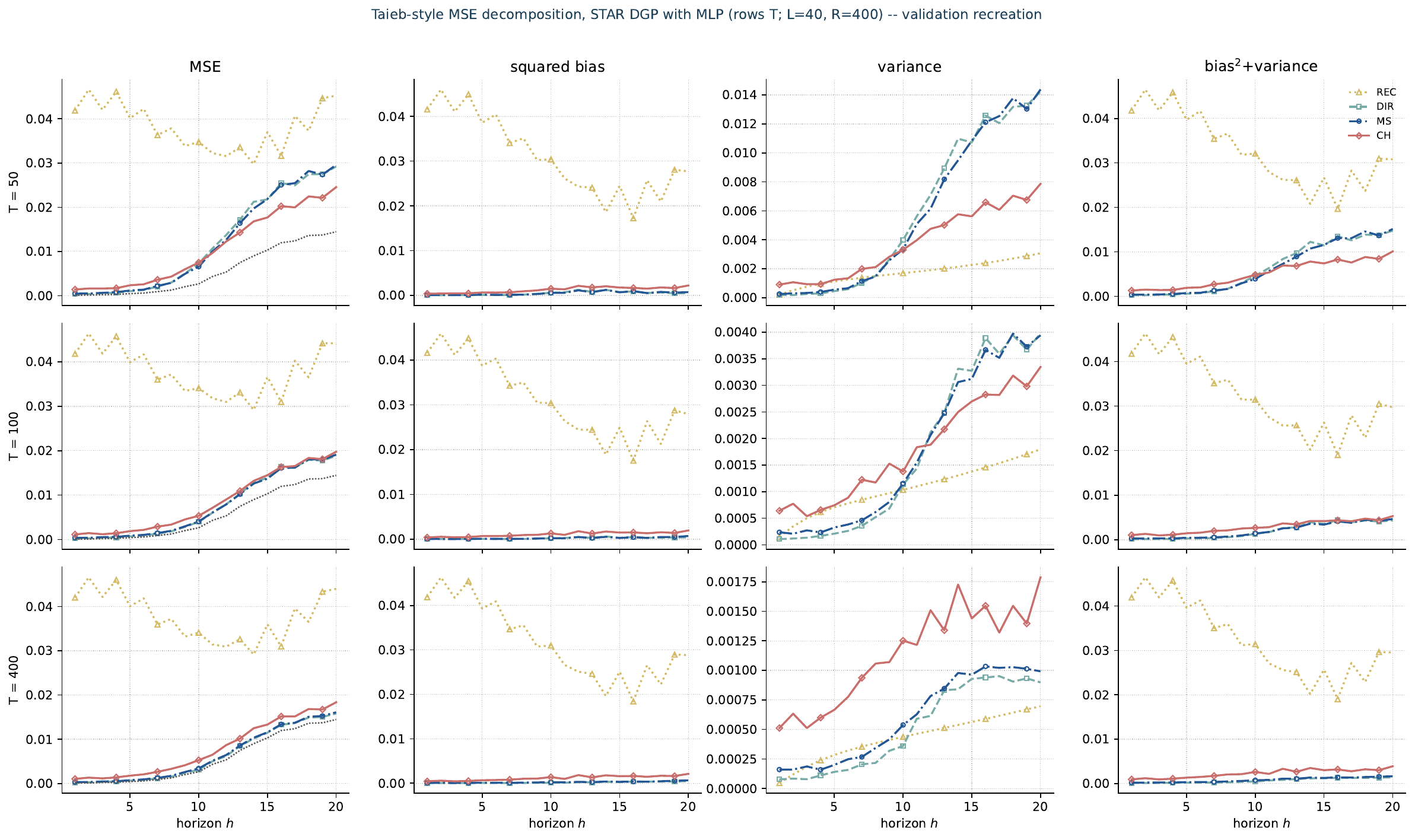}
    \caption{MSE decomposition on the STAR process~\citep{taieb2015bias}, in the same layout as Figure~\ref{fig:taieb_nlar}. The recurrent strategy exhibits visible bias from rollout limitations on the smooth-transition nonlinearity, whereas all other strategies—including \ourscomplete (\ours)—are near-optimal on this nearly deterministic process.}
    \label{fig:taieb_star}
\end{figure}
 \clearpage
\section{Unseen Horizons Experiment Details}
\label{appendix:interpolation_extrapolation}

\textbf{Data.} The experiment in Section~\ref{sec:05_interp_extrap} uses zero-mean Gaussian processes. A Gaussian process has a value at each real time, so it gives true targets at fractional and unseen horizons. We use three kernels, with $\tau = t - t'$:
\begin{equation*}
\begin{aligned}
K_{\mathrm{SE}}(\tau) &= \exp\!\Big(-\frac{\tau^2}{2\ell_{\mathrm{se}}^2}\Big), \qquad\qquad
K_{\mathrm{Per}}(\tau) = \exp\!\Big(-\frac{2\sin^2(\pi|\tau|/p)}{\ell^2}\Big), \; \\
K_{\mathrm{LP}}(\tau) &= K_{\mathrm{Per}}(\tau)\,K_{\mathrm{SE}}(\tau).
\end{aligned}
\end{equation*}
Table~\ref{table:gp_regimes} gives the kernel and the parameters of each regime. We add observation noise with $\sigma_n = 0.5$ to each series.

\begin{table}[ht]
\centering
\caption{Kernels and parameters of the five Gaussian-process regimes.}
\label{table:gp_regimes}
\small
\begin{tabular}{@{}llccc@{}}
\toprule
Regime & Kernel & $\ell$ & $p$ & $\ell_{\mathrm{se}}$ \\
\midrule
Mean-reverting & $K_{\mathrm{LP}}$  & 3  & 12 & 12 \\
Aperiodic      & $K_{\mathrm{SE}}$  & -- & -- & 16 \\
Periodic       & $K_{\mathrm{Per}}$ & 3  & 12 & -- \\
QP (slow)      & $K_{\mathrm{LP}}$  & 3  & 12 & 48 \\
QP (fast)      & $K_{\mathrm{LP}}$  & 3  & 12 & 24 \\
\bottomrule
\end{tabular}
\end{table}

\textbf{Exact Conditional Mean.} For a context $\mathbf{y}_{:t}$, the posterior Gaussian-process gives the exact conditional mean $\mu_{t+h|t}$ at each real horizon $h$. The posterior includes the same observation noise, so the regret of Eq.~\eqref{eq:gp_regret} is exact. The forecasters do not know the kernel.

\textbf{Models.} All strategies use the same two-layer encoder in a context window of 16 observations. The \ours\ decoder is the branch--trunk expansion of Eq.~\eqref{eq:ch_concat}. \MS\ uses one shared output layer for all horizons, \DIR\ uses one output layer for each horizon, and \REC\ applies a one-step output layer recursively. \MS\ and \ours\ have approximately 5,000 parameters each.

\textbf{Training and evaluation.} Each training sample has 64 series. We train each strategy for 8,000 steps on the integer horizons $h \le 12$. We repeat the training on $R = 40$ independent samples. All strategies use pinball loss on quantile levels $\{0.05, 0.1, 0.25, 0.5, 0.75, 0.9, 0.95\}$, and the point forecast is the median. For a Gaussian posterior, the median is equal to the conditional mean $\mu_{t+h|t}$. We evaluate the integer horizons seen, the half-integer horizons, and the integer horizons $12 < h \le 24$.
 \clearpage
\section{Qualitative Forecast Exploration}
\label{appendix:qualitative_forecasts}

\begin{figure}[ht!]
    \centering
    \includegraphics[width=\linewidth]{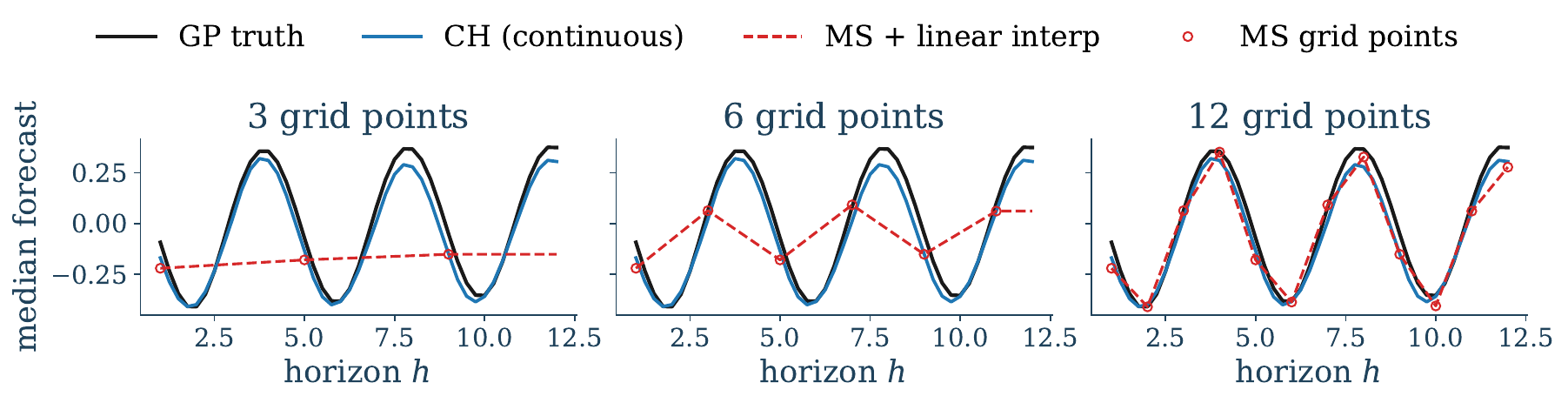}
    \caption{Forecasts of an oscillating Gaussian Process signal. Black: GP truth. Blue: \ours. Red dashed: \MS\ with linear interpolation, with markers at its output horizons. Left to right: coarse, intermediate, and full \MS\ grid.}
    \label{fig:interp_grid_density}
\end{figure}

Here we complement the accuracy results of Section~\ref{sec:05_interp_extrap}, with a qualitative exploration of the outputs of discrete and \ours forecast strategy.

The strategies \MS, \REC, \DIR produce forecasts discretely with the output grid $\{0,1,\dots,H\}$. For any other horizon $h \in [0,H]$, to obtain the forecast, we must interpolate between two adjacent horizons. 

Horizons therefore in the interpolation regime can be thought of as interpolant knots. The forecast accuracy between the knots depends on the distance $\Delta$ between them. For an output grid with $n$ intervals on $[0,H]$, $\Delta = H/n$.

For linear interpolation, the error between two knots is not more than $\Delta^{2}/8$ times the maximum curvature $\max_h |f''(h \mid \mathbf{y}_{:t})|$ of the true forecast. Thus, if the grid has half the number of horizons, $\Delta$ doubles, and the error bound becomes four times larger~\citep{deboor1978practical}. A discrete forecast strategy can show an oscillation only if the distance between knots is less than half of the target signal period~\citep{shannon1949communication}. If the horizon distance is larger, the discrete strategies cannot contain the cycles. Then, no interpolation method can find the cycle again, and the result is aliasing. The coarse panel of Figure~\ref{fig:interp_grid_density} shows this result: the \MS\ interpolant is an almost straight line.

The strategy \ours\ bypasses discrete horizons. It learns the forecast as a continuous function of $h$, so the interpolation error of the \MS\ grid does not occur. But \ours\ also trains only on integer horizons. Its accuracy between these horizons depends on the smoothness of the learned horizon basis functions $\bphi_k(h)$, not on the density of the grid.

 \clearpage
\section{Horizon-Basis Ablation}
\label{appendix:basis_ablation}

The horizon basis $\hat\bphi(h)$ lets \ours\ forecast at unseen horizons. All variants use the same encoder, parameter budget, training schedule, and branch coefficients. Only $\hat\bphi(h)$ changes.

\textbf{Protocol.} We use the Gaussian-process data of \ref{appendix:interpolation_extrapolation}, with 48 series in each training sample, 3,000 training steps, and $R=24$ replicates. Some kernel parameters are different from Table~\ref{table:gp_regimes}. The mean-reverting regime uses $K_{\mathrm{SE}}$ with $\ell_{\mathrm{se}}=3$, and the periodic regime uses $\ell=1.5$. The slow and the fast quasi-periodic regimes use $\ell_{\mathrm{se}}=24$ and $\ell_{\mathrm{se}}=8$. The metric is the RMSE of Eq.~\eqref{eq:rmse}.

\textbf{Basis blocks.} The basis uses five blocks. Each block is a function of the horizon $h$, or a constant:
\begin{align}
\mathbf{b}_{\mathrm{lin}}(h) &= h/H, &
\mathbf{b}_{\mathrm{quad}}(h) &= (h/H)^2, \nonumber\\
\mathbf{b}_{\mathrm{four}}(h) &= [\, \sin(\omega_k h),\, \cos(\omega_k h) \,]_{k=1}^{m}, &
\mathbf{b}_{\mathrm{tbl}}(h) &= (1-\lambda)\,\mathbf{E}^{(\mathrm{tbl})}_{\lfloor h\rfloor} + \lambda\,\mathbf{E}^{(\mathrm{tbl})}_{\lceil h\rceil}, \nonumber\\
\mathbf{b}_{\mathrm{freq}} &= \mathbf{E}^{(\mathrm{freq})}. && \nonumber
\end{align}
The blocks $\mathbf{b}_{\mathrm{lin}}$ and $\mathbf{b}_{\mathrm{quad}}$ give a polynomial trend. The Fourier block $\mathbf{b}_{\mathrm{four}}$ has fixed or learnable frequencies $\omega_k$. The table block $\mathbf{b}_{\mathrm{tbl}}$ interpolates a learned embedding table $\mathbf{E}^{(\mathrm{tbl})}\in\mathbb{R}^{H\times d_t}$, with $\lambda = h-\lfloor h\rfloor$. For $h>H$, it keeps the last value $\mathbf{E}^{(\mathrm{tbl})}_H$. The block $\mathbf{b}_{\mathrm{freq}}$ is a learned frequency embedding, and it does not depend on $h$.

\textbf{Composite basis.} The basis is a linear combination of the blocks:
\begin{equation}
\hat\bphi(h) = \sum_{b \in \mathcal{B}} \mathbf{A}_b\, \mathbf{b}_b(h),
\qquad \mathcal{B} = \{\mathrm{lin}, \mathrm{quad}, \mathrm{four}, \mathrm{tbl}, \mathrm{freq}\}.
\label{eq:basis_composite}
\end{equation}
Each matrix $\mathbf{A}_b \in \mathbb{R}^{K\times d_b}$ is learned, with $d_b$ the dimension of block $b$. A variant removes block $b$ with $\mathbf{A}_b = 0$. The basis does not depend on the history, so it is separable, as in Eq.~\eqref{eq:ch_concat}.

\textbf{Non-separable variants.} Two variants use the encoder state $\hat\btheta_t$ in the basis. The \emph{gated} variant multiplies the quadratic and the Fourier blocks by gates from the history, with $\mathbf{A}_{\mathrm{freq}} = 0$:
\begin{equation}
\begin{split}
\hat\bphi(h) = \mathbf{A}_{\mathrm{lin}}\, \mathbf{b}_{\mathrm{lin}}(h)
 + \gamma_{\mathrm{quad}}\, \mathbf{A}_{\mathrm{quad}}\, \mathbf{b}_{\mathrm{quad}}(h)
 + \gamma_{\mathrm{four}}\, \mathbf{A}_{\mathrm{four}}\, \mathbf{b}_{\mathrm{four}}(h)
 + \mathbf{A}_{\mathrm{tbl}}\, \mathbf{b}_{\mathrm{tbl}}(h), \\
\qquad (\gamma_{\mathrm{quad}},\, \gamma_{\mathrm{four}}) = \sigma\big(\mathrm{gate}(\hat\btheta_t)\big).
\end{split}
\label{eq:basis_gate}
\end{equation}
The \emph{per-series} variant uses the frequency embedding ($\mathbf{A}_{\mathrm{freq}} \neq 0$) and replaces $\mathbf{E}^{(\mathrm{freq})}$ with $\mathbf{E}^{(\mathrm{freq})}(\hat\btheta_t)$, which the encoder gives. In both variants, the basis depends on the history, so it is not separable.

\textbf{Ablations.} Each ablation changes one part of Eq.~\eqref{eq:basis_composite} or Eq.~\eqref{eq:basis_gate}. The anchor uses all blocks except the frequency embedding ($\mathbf{A}_{\mathrm{freq}} = 0$), with learnable $\omega_k$ and $K=16$. The ablations are:
\begin{itemize}
    \item Composition (Table~\ref{table:basis_ablation_composition}): all seven combinations of the polynomial (lin and quad), the Fourier, and the table blocks, with $\mathbf{A}_{\mathrm{freq}} = 0$ and $\mathbf{A}_b = 0$ for each unused block.
    \item Basis Dimensionality (Table~\ref{table:basis_ablation_k}): $K \in \{2, 4, 8, 16, 32\}$.
    \item Frequency representation (Table~\ref{table:basis_ablation_freqsep}): fixed or learnable $\omega_k$, the frequency embedding with and without the Fourier block ($\mathbf{A}_{\mathrm{four}} = 0$), the per-series embedding, and the gate.
\end{itemize}
In Table~\ref{table:basis_ablation_freqsep}, the \emph{Sep.} column shows the separable variants, to which \ref{appendix:universal_approximation} and \ref{appendix:ch_decomposition} apply. \MS\ and \DIR\ are not in these tables, because they cannot forecast at unseen horizons.

\textbf{Results.} On the seen and the interpolation horizons, all variants have almost the same error. The exceptions are the polynomial-only basis and $K=2$ on the periodic regime. On the extrapolation horizons, the variants are different:
\begin{itemize}
    \item No composition is the best in all regimes. The table-only basis is the best in three regimes.
    \item The polynomial-only basis has the highest error in three regimes.
    \item From $K=4$ to $K=32$, the error marginally decreases by 5\% or less.
    \item Without the Fourier block, the frequency embedding has a higher error on the periodic regimes.
    \item The non-separable variants have the lowest error in four regimes. Their advantage over the best separable variant is only 1.3\% or less.
\end{itemize}

\begin{table}[t!]
\caption{Ablation of the basis composition. RMSE on the seen, interpolation, and extrapolation horizons of the five regimes. The anchor is the full composite basis with learnable frequencies and $K=16$. Bold shows the lowest extrapolation error.}
\label{table:basis_ablation_composition}
\centering\scriptsize
\begin{tabular}{ll|ccc}
\toprule
Regime & Basis composition & Seen & Interp. & Extrap. \\
\midrule
\multirow{7}{*}{Mean-reverting}
  & polynomial only            & 1.092 & 1.100 & 1.610 \\
  & Fourier only               & 1.092 & 1.101 & 1.226 \\
  & table only                 & 1.091 & 1.099 & \textbf{1.153} \\
  & polynomial $+$ Fourier     & 1.094 & 1.103 & 1.263 \\
  & polynomial $+$ table       & 1.091 & 1.099 & 1.333 \\
  & Fourier $+$ table          & 1.091 & 1.099 & 1.188 \\
  & full composite (anchor)    & 1.092 & 1.101 & 1.217 \\
\midrule
\multirow{7}{*}{Aperiodic}
  & polynomial only            & 0.715 & 0.713 & 1.077 \\
  & Fourier only               & 0.715 & 0.713 & 1.085 \\
  & table only                 & 0.715 & 0.712 & \textbf{1.030} \\
  & polynomial $+$ Fourier     & 0.715 & 0.713 & 1.057 \\
  & polynomial $+$ table       & 0.715 & 0.712 & 1.054 \\
  & Fourier $+$ table          & 0.715 & 0.713 & 1.065 \\
  & full composite (anchor)    & 0.715 & 0.712 & 1.054 \\
\midrule
\multirow{7}{*}{Periodic}
  & polynomial only            & 0.674 & 0.658 & 5.339 \\
  & Fourier only               & 0.643 & 0.638 & 0.740 \\
  & table only                 & 0.642 & 0.633 & 0.925 \\
  & polynomial $+$ Fourier     & 0.642 & 0.637 & 0.762 \\
  & polynomial $+$ table       & 0.642 & 0.633 & 1.138 \\
  & Fourier $+$ table          & 0.642 & 0.636 & \textbf{0.691} \\
  & full composite (anchor)    & 0.641 & 0.635 & 0.729 \\
\midrule
\multirow{7}{*}{Quasi-periodic (slow)}
  & polynomial only            & 0.700 & 0.698 & 1.391 \\
  & Fourier only               & 0.683 & 0.683 & 0.943 \\
  & table only                 & 0.690 & 0.688 & 0.935 \\
  & polynomial $+$ Fourier     & 0.683 & 0.683 & 0.929 \\
  & polynomial $+$ table       & 0.691 & 0.689 & 0.989 \\
  & Fourier $+$ table          & 0.683 & 0.683 & 0.926 \\
  & full composite (anchor)    & 0.683 & 0.682 & \textbf{0.919} \\
\midrule
\multirow{7}{*}{Quasi-periodic (fast)}
  & polynomial only            & 0.983 & 0.984 & 1.228 \\
  & Fourier only               & 0.981 & 0.983 & 1.280 \\
  & table only                 & 0.981 & 0.982 & \textbf{1.175} \\
  & polynomial $+$ Fourier     & 0.982 & 0.984 & 1.210 \\
  & polynomial $+$ table       & 0.981 & 0.983 & 1.218 \\
  & Fourier $+$ table          & 0.980 & 0.982 & 1.242 \\
  & full composite (anchor)    & 0.982 & 0.983 & 1.202 \\
\bottomrule
\end{tabular}
\end{table}

\begin{table}[t!]
\caption{Ablation of the number $K$ of basis functions, with the full composite basis. RMSE on the seen, interpolation, and extrapolation horizons of the five regimes.}
\label{table:basis_ablation_k}
\centering\scriptsize
\begin{tabular}{ll|ccc}
\toprule
Regime & $K$ & Seen & Interp. & Extrap. \\
\midrule
\multirow{5}{*}{Mean-reverting}
  & $K=2$ & 1.091 & 1.102 & 1.267 \\
  & $K=4$ & 1.093 & 1.103 & 1.234 \\
  & $K=8$ & 1.093 & 1.102 & 1.229 \\
  & $K=16$ & 1.092 & 1.101 & 1.217 \\
  & $K=32$ & 1.092 & 1.100 & 1.203 \\
\midrule
\multirow{5}{*}{Aperiodic}
  & $K=2$ & 0.716 & 0.716 & 1.059 \\
  & $K=4$ & 0.715 & 0.713 & 1.054 \\
  & $K=8$ & 0.715 & 0.713 & 1.055 \\
  & $K=16$ & 0.715 & 0.712 & 1.054 \\
  & $K=32$ & 0.715 & 0.712 & 1.052 \\
\midrule
\multirow{5}{*}{Periodic}
  & $K=2$ & 0.665 & 0.704 & 1.529 \\
  & $K=4$ & 0.641 & 0.640 & 0.748 \\
  & $K=8$ & 0.641 & 0.638 & 0.740 \\
  & $K=16$ & 0.641 & 0.635 & 0.729 \\
  & $K=32$ & 0.643 & 0.636 & 0.713 \\
\midrule
\multirow{5}{*}{Quasi-periodic (slow)}
  & $K=2$ & 0.699 & 0.701 & 0.971 \\
  & $K=4$ & 0.684 & 0.684 & 0.934 \\
  & $K=8$ & 0.683 & 0.683 & 0.930 \\
  & $K=16$ & 0.683 & 0.682 & 0.919 \\
  & $K=32$ & 0.683 & 0.682 & 0.915 \\
\midrule
\multirow{5}{*}{Quasi-periodic (fast)}
  & $K=2$ & 0.982 & 0.987 & 1.241 \\
  & $K=4$ & 0.981 & 0.983 & 1.213 \\
  & $K=8$ & 0.981 & 0.983 & 1.201 \\
  & $K=16$ & 0.982 & 0.983 & 1.202 \\
  & $K=32$ & 0.981 & 0.982 & 1.199 \\
\bottomrule
\end{tabular}
\end{table}

\begin{table}[t!]
\caption{Ablation of the frequency representation. RMSE on the seen, interpolation, and extrapolation horizons of the five regimes. A check mark in the Sep.\ column shows a separable basis. Bold shows the lowest extrapolation error in each regime.}
\label{table:basis_ablation_freqsep}
\centering\scriptsize
\begin{tabular}{ll|ccc|c}
\toprule
Regime & Frequency representation & Seen & Interp. & Extrap. & Sep. \\
\midrule
\multirow{6}{*}{Mean-reverting}
  & fixed Fourier dictionary         & 1.092 & 1.101 & 1.234 & \cmark \\
  & learnable Fourier (anchor)       & 1.092 & 1.101 & 1.217 & \cmark \\
  & frequency embedding, shared         & 1.093 & 1.102 & 1.213 & \cmark \\
  & frequency embedding, per-series         & 1.092 & 1.100 & \textbf{1.204} & \xmark \\
  & frequency embedding only (no Fourier)   & 1.091 & 1.099 & 1.312 & \cmark \\
  & gated composite                  & 1.092 & 1.100 & 1.212 & \xmark \\
\midrule
\multirow{6}{*}{Aperiodic}
  & fixed Fourier dictionary         & 0.715 & 0.713 & 1.047 & \cmark \\
  & learnable Fourier (anchor)       & 0.715 & 0.712 & 1.054 & \cmark \\
  & frequency embedding, shared         & 0.715 & 0.713 & 1.049 & \cmark \\
  & frequency embedding, per-series         & 0.716 & 0.713 & 1.052 & \xmark \\
  & frequency embedding only (no Fourier)   & 0.715 & 0.712 & 1.046 & \cmark \\
  & gated composite                  & 0.715 & 0.713 & \textbf{1.036} & \xmark \\
\midrule
\multirow{6}{*}{Periodic}
  & fixed Fourier dictionary         & 0.642 & 0.637 & 0.746 & \cmark \\
  & learnable Fourier (anchor)       & 0.641 & 0.635 & 0.729 & \cmark \\
  & frequency embedding, shared         & 0.642 & 0.635 & 0.703 & \cmark \\
  & frequency embedding, per-series         & 0.642 & 0.635 & \textbf{0.701} & \xmark \\
  & frequency embedding only (no Fourier)   & 0.642 & 0.633 & 1.098 & \cmark \\
  & gated composite                  & 0.642 & 0.635 & 0.733 & \xmark \\
\midrule
\multirow{6}{*}{Quasi-periodic (slow)}
  & fixed Fourier dictionary         & 0.683 & 0.683 & 0.917 & \cmark \\
  & learnable Fourier (anchor)       & 0.683 & 0.682 & 0.919 & \cmark \\
  & frequency embedding, shared         & 0.683 & 0.682 & \textbf{0.914} & \cmark \\
  & frequency embedding, per-series         & 0.684 & 0.683 & 0.918 & \xmark \\
  & frequency embedding only (no Fourier)   & 0.690 & 0.688 & 0.981 & \cmark \\
  & gated composite                  & 0.685 & 0.684 & 0.938 & \xmark \\
\midrule
\multirow{6}{*}{Quasi-periodic (fast)}
  & fixed Fourier dictionary         & 0.981 & 0.983 & 1.195 & \cmark \\
  & learnable Fourier (anchor)       & 0.982 & 0.983 & 1.202 & \cmark \\
  & frequency embedding, shared         & 0.981 & 0.983 & 1.199 & \cmark \\
  & frequency embedding, per-series         & 0.980 & 0.982 & 1.200 & \xmark \\
  & frequency embedding only (no Fourier)   & 0.981 & 0.983 & 1.205 & \cmark \\
  & gated composite                  & 0.982 & 0.983 & \textbf{1.180} & \xmark \\
\bottomrule
\end{tabular}
\end{table}

 \clearpage

\section{Dataset Details}
\label{appendix:dataset_details}

\begin{table}[!ht]
\caption{Summary of forecasting datasets used in our main empirical study.} \label{table:datasets}
\centering
\footnotesize
\resizebox{0.9\textwidth}{!}{
\begin{tabular}{llcccccc}
\toprule
                         & Frequency & Seasonality & Horizon                     & Series                     & Min Length                     & Max Length                     & \% Erratic                     \\ \midrule
\multirow{3}{*}{M1}      & Monthly   & 12          & 18                          & 617                        & 48                             & 150                            & 0                              \\
                         & Quarterly & 4           & 8                           & 203                        & 18                             & 114                            & 0                              \\
                         & Yearly    & 1           & 6                           & 181                        & 15                             & 58                             & 0                              \\ \midrule
\multirow{4}{*}{M3}      & Other     & 4           & 8                           & 174                        & 71                             & 104                            & 0                              \\
                         & Monthly   & 12          & 18                          & 1428                       & 66                             & 144                            & 2                              \\
                         & Quarterly & 4           & 8                           & 756                        & 24                             & 72                             & 1                              \\
                         & Yearly    & 1           & 6                           & 645                        & 20                             & 47                             & 10                             \\ \midrule
\multirow{6}{*}{M4}      & Hourly    & 24          & 48                          & 414                        & 748                            & 1008                           & 17                             \\ 
                         & Daily     & 1           & 14                          & 4,227                      & 107                            & 9933                           & 2                              \\
                         & Weekly    & 1           & 13                          & 359                        & 93                             & 2610                           & 16                             \\
                         & Monthly   & 12          & 18                          & 48,000                     & 60                             & 2812                           & 6                              \\
                         & Quarterly & 4           & 8                           & 24,000                     & 24                             & 874                            & 11                             \\
                         & Yearly    & 1           & 6                           & 23,000                     & 19                             & 841                            & 18                             \\ \midrule
\multirow{3}{*}{Tourism} & Monthly   & 12          & 18                          & 366                        & 91                             & 333                            & 51                             \\
                         & Quarterly & 4           & 8                           & 427                        & 30                             & 130                            & 39                             \\
                         & Yearly    & 1           & 6                           & 518                        & 11                             & 47                             & 23                             \\ \bottomrule
\end{tabular}
}
\end{table}

\textbf{M1.} The \Mone competition~\citep{makridakis1982m1_competition} has 1,001 monthly, quarterly, and yearly series from demography, industry, and economics. In this competition, simple methods such as \ETS~\citep{holt1957exponential_smoothing} were frequently more accurate than complex methods. The \Mone competition initiated a research legacy that emphasized forecast accuracy.

\textbf{M3.} The \Mthree competition~\citep{makridakis2000m3_competition} has 3,003 series of business, demography, finance, and economics. All values are positive, and no series is intermittent~\citep{syntetos2005syntetos_categorization}. The \THeta\ method~\citep{hyndman2003unmasking_theta} had the best overall accuracy.

\textbf{M4.} The \Mfour competition~\citep{makridakis2020m4_competition} has 100,000 series at six frequencies, from demography, finance, industry, and economics. It added the evaluation of prediction intervals. \ESRNN~\citep{smyl2019esrnn}, a hybrid of exponential smoothing and a neural network, won the competition. \Mfour\ also made cross-learning and global models popular~\citep{spiliotis2021cross_learning}.

\textbf{Tourism.} The \Tourism dataset~\citep{athanosopoulus2011tourism_competition} has 1,311 monthly, quarterly, and yearly series of tourism demand. This competition was the first to use the mean absolute scaled error (MASE). Of all series, 36\% are erratic or intermittent. At the yearly frequency, few methods were more accurate than the Naïve1 method.
 \clearpage
\section{Training Methodology and Hyperparameters}
\label{appendix:training_methodology}

\subsection{Network Hyperparameters}
\label{appendix:hyperparameters}

We train \MQForecaster varying encoders between \LSTM, \CNN, \Transformer, and \Sfour modules. We train models with hyperparameter tuning using Optuna. Using 5 trials, we train individual models and select the hyperparameters that minimize the validation RMSE. Training times mostly depend on the architecture; however, we restrict the SGD training steps to 45K per architecture. Tables~\ref{table:shared_encoder_hyperparameters} and ~\ref{table:hyperparameters_lstm_encoder}--\ref{table:hyperparameters_s4_encoder} list the shared hyperparameters and encoder-specific hyperparameters, respectively. 

\begin{table}[!ht]
    \centering
    \caption{Shared hyperparameters across encoder types}
    \label{table:shared_encoder_hyperparameters}
    \resizebox{0.47\textwidth}{!}{
    \begin{tabular}{lc}
        \toprule
        \textsc{Hyperparameter}                                         & \thead{ \textsc{Values}}  \\ \midrule
        Single GPU SGD Batch Size\textsuperscript{*}                   & \{4, 8, 16\}                         \\
        Initial learning rate                                          & loguniform(1e-4, 5e-2)                    \\
        Maximum Training steps $S_{max}$                               & 45,000                    \\ 
        Learning rate decay                                            & 0.1                       \\
        Learning rate step size                                        & 15,000                    \\
        Scaler type                                                   & Standard                  \\ 
        H-Agnostic Decoder Dimension                                   & 100                       \\
        H-Specific Decoder Dimension                                   & 20                        \\
        \bottomrule
        \end{tabular}
        }
    \centering
\end{table}

\begin{table}[!ht]
    \begin{minipage}{.5\linewidth}
      \caption{\LSTM} \label{table:hyperparameters_lstm_encoder}
      \centering
      \tiny
        \begin{tabular}{lc}
        \toprule
        \textsc{Hyperparameter}                                         & \thead{ \textsc{Values}}  \\ \midrule
        Main Activation Functions & Sigmoid, tanh  \\
        Encoder Dimension                                              & \{128, 256\}  \\
        \multirow{2}{*}{Dilations} & \{[[1,2], [4,8]], \\
        {} & [[1,2], [4,8], [16,32]]\} \\
        \bottomrule
        \end{tabular}
    \end{minipage}%
\hspace{1.cm}
    \begin{minipage}{.5\linewidth}
      \caption{\CNN} \label{table:hyperparameters_cnn_encoder}
      \centering
      \tiny
        \begin{tabular}{lc}
        \toprule
        \textsc{Hyperparameter}                                         & \thead{ \textsc{Values}}  \\ \midrule
        Main Activation Function                                        & {ReLU}                    \\
        Encoder Dimension & \{128, 256\} \\
        Temporal Convolution Kernel Size                                & \{2, 5\}                       \\
        \multirow{2}{*}{Temporal Convolution Dilations}                & \{[1, 2, 4, 8], \\
        {} & [1, 2, 4, 8, 16, 32]\}       \\
        \bottomrule
        \end{tabular}
    \end{minipage}%
\end{table}

\begin{table}[!ht]
    \begin{minipage}{.5\linewidth}
      \caption{\Transformer} \label{table:hyperparameters_transformer_encoder}
      \centering
      \tiny
        \begin{tabular}{lc}
        \toprule
        \textsc{Hyperparameter}                                         & \thead{ \textsc{Values}}  \\ \midrule
        Main Activation Functions & Sigmoid, tanh  \\
        Encoder Dimension                                            & \{128, 256\}                       \\ 
        Projection Dimension                                            & \{128, 256\}                       \\ 
        Input Patch Lengths                                            & \{1, 8, 32, 96\} \\
        Attention Dilations                                            & \{[2, 6, 8], [2, 4, 8, 16]\} \\
        Attention Dropout                                              & 0.1                       \\
        Number of Attention Heads                                      & 4                         \\
        \bottomrule
        \end{tabular}
    \end{minipage}%
\hspace{1.cm}
    \begin{minipage}{.5\linewidth}
      \caption{State Space \Sfour} \label{table:hyperparameters_s4_encoder}
      \centering
      \tiny
        \begin{tabular}{lc}
        \toprule
        \textsc{Hyperparameter}                                         & \thead{ \textsc{Values}}  \\ \midrule
        Main Activation Function                                        & {GELU}                    \\
        Encoder Dimension                                            & \{128, 256\}                       \\ 
        Projection Hidden Size                                            & \{128, 256\}              \\ 
        \bottomrule
        \end{tabular}
    \end{minipage}%
\end{table}

\subsection{Parameter-Matched Strategy Comparison}
\label{appendix:param_matching}

We compare the strategies at the same number of parameters, so that an accuracy difference comes from the horizon strategy, not from the model size.

The decoders of \MS\ and \ours\ scale differently with the horizon $H$. The \MS\ decoder has one output block for each horizon, so its size increases with $H$. The \ours\ decoder uses a basis $\bphi(h)$ with a fixed size. Without matching, \MS\ has more parameters than \ours, and the difference increases with $H$.

To match parameters, we marginally decrease the width of the encoder output for \MS, so that the totals agree. For example, the width is 32 for \ours\ and 19 for \MS\ with the \CNN\ encoder. Then the totals are different by 3.1\% or less (Table~\ref{table:param_counts}). We count the parameters directly, because the same width gives different counts for different decoders.

\begin{table}[ht!]
\caption{Trainable \MQForecaster parameters for each encoder and strategy. A dash marks a strategy not trained with that encoder; counts are functional (CH basis/embeddings disabled for \DIR and MS). \ours, MS, and \REC are single global models. $^{*}$\DIR trains one model \emph{per horizon}; the listed count is per horizon, split as encoder~$+$~decoder (CNN: $8{,}006+1{,}174$). The encoder matches the global model; the decoder is $1/H$ of the global decoder, so over $H{=}18$ horizons the \DIR decoders sum to $\approx$ one global decoder ($18\times1{,}174{=}21{,}132$, vs.\ $19{,}221$) while the encoder is replicated ($18\times8{,}006$).}
\label{table:param_counts}
\centering
\scriptsize
\resizebox{\linewidth}{!}{
\begin{tabular}{l|cccc|cccc|cccc|cccc}
\toprule
& \multicolumn{4}{c|}{\LSTM}
& \multicolumn{4}{c|}{\CNN}
& \multicolumn{4}{c|}{\Transformer}
& \multicolumn{4}{c}{\Sfour} \\
Model
& \ours & MS & \DIR{}$^{*}$ & \REC
& \ours & MS & \DIR{}$^{*}$ & \REC
& \ours & MS & \DIR{}$^{*}$ & \REC
& \ours & MS & \DIR{}$^{*}$ & \REC \\
\midrule
Global model & 78{,}482 & 77{,}865 & 7{,}178 & 78{,}482 & 63{,}154 & 65{,}125 & 9{,}477 & 63{,}154 & 81{,}746 & 80{,}868 & 8{,}641 & 81{,}746 & 85{,}714 & 86{,}674 & 13{,}961 & 85{,}714 \\
\bottomrule
\end{tabular}}
\end{table}

\subsection{Regret-Based Dataset Sampling for Cross-Learning}
\label{appendix:regret_sampler}

We train one global model on the 16 datasets of Section~\ref{sec:06_evaluation}, by cross-learning~\citep{spiliotis2021cross_learning}. The datasets are very different in size, so proportional sampling gives the small datasets few gradient updates. Thus, we change the sampling probability of each dataset during training. A dataset gets a larger probability when its validation error is high relative to a reference for that dataset. This method is a variant of group distributionally robust optimization~\citep{sagawa2020groupdro} and of data-mixture reweighting of \citet{xie2023doremi}.

\clearpage
\subsection{The Continuous-Horizon Learning Objective}
\label{appendix:ch_objective}

\textbf{\MS\ Learning Objective.} Theorem~\ref{thm:ch_dominates_body} shows the benefit of training on fractional horizons. But the M-series experiments compare the inductive bias of the decoders, with the learning objective fixed. Thus, we train \ours\ on the integer horizons $h\in\{1,\dots,H\}$, with the traditional \MS\ objective. The uniform measure $\tfrac{1}{H}\,\mathrm{d}h$ of Eq.~\eqref{eq:erm} then becomes the uniform measure on the integer grid:
\begin{equation}
    \hat{\mathbf{w}} := \argmin_{\mathbf{w}}\;
    \frac{1}{T \times H}\sum_{t=1}^{T}\sum_{h=1}^{H}
    L\big(y_{t+h},\, g(h \mid \mathbf{y}_{:t}, \mathbf{w})\big).
    \label{eq:ch_objective_grid}
\end{equation}
The loss $L$ is the MSE. \MS\ minimizes Eq.~\eqref{eq:ch_objective_grid} with all $H$ horizons in each training step. For \ours, we used the same loss and the same pooled data, but we sampled the horizons.

\textbf{\ours\ Learning Objective.} In each training step, we sample a random subset $\mathcal{S} \subseteq \{1,\dots,H\}$ of the horizons for the full mini-batch. We drop each horizon independently with probability 0.2 and we keep at least 10 horizons. Each horizon has the same probability to be in $\mathcal{S}$, so
\begin{equation}
    \mathbb{E}_{\mathcal{S}}\!\left[\frac{1}{|\mathcal{S}|}\sum_{h \in \mathcal{S}} L\big(y_{t+h},\, g(h \mid \mathbf{y}_{:t}, \mathbf{w})\big)\right]
    = \frac{1}{H}\sum_{h=1}^{H} L\big(y_{t+h},\, g(h \mid \mathbf{y}_{:t}, \mathbf{w})\big).
    \label{eq:horizon_sampling}
\end{equation}
Thus, a training step with $\mathcal{S}$ minimizes Eq.~\eqref{eq:ch_objective_grid} in expectation. In our experiments, the number of sampled horizons changed the validation accuracy by less than 2\%.

 \clearpage
\section{Additional Point Forecast Metrics}
\label{appendix:smape_mase}

We complement the relAvgRMSE of Table~\ref{table:main_mse} with the two absolute-error metrics most used in the M-competitions, so that our results can be read against previous work on these datasets. Both metrics use the mean forecast $\hat{y}_{t+h}$ and the pool over all series $i$, the forecast creation dates $t \in [0,1,...,T]$ and the horizons $h$ of the temporal cross-validation of Section~\ref{sec:06_evaluation}. 

\begin{table}[h!]
\caption{Empirical evaluation of point forecasts. Symmetric Mean Absolute Percentage Error (sMAPE, Eq.~\eqref{eq:smape}), averaged over 5 runs. Lower measurements are preferred. The methods without standard deviation have deterministic solutions. For each encoder we compare four forecasting strategies: our \ourscomplete (\ours), multi-step (MS), direct (\DIR), and recurrent (\REC).}

\label{table:main_smape}
\centering
\scriptsize
\resizebox{\textwidth}{!}{
\begin{tabular}{ll|cccc|cccc|cccc|cccc|cc}
\toprule
& & \multicolumn{4}{c|}{\LSTM}
& \multicolumn{4}{c|}{\CNN}
& \multicolumn{4}{c|}{\Transformer}
& \multicolumn{4}{c|}{\Sfour}
& \multicolumn{2}{c}{StatsForecast} \\
& Freq 
& \multicolumn{1}{c}{\ours} & \multicolumn{1}{c}{MS} & \multicolumn{1}{c}{\DIR} & \multicolumn{1}{c|}{\REC} 
& \multicolumn{1}{c}{\ours} & \multicolumn{1}{c}{MS} & \multicolumn{1}{c}{\DIR} & \multicolumn{1}{c|}{\REC}
& \multicolumn{1}{c}{\ours} & \multicolumn{1}{c}{MS} & \multicolumn{1}{c}{\DIR} & \multicolumn{1}{c|}{\REC}
& \multicolumn{1}{c}{\ours} & \multicolumn{1}{c}{MS} & \multicolumn{1}{c}{\DIR} & \multicolumn{1}{c|}{\REC} 
& \ETS & \ARIMA \\
\midrule
\xdef\prevdataset{}
\begin{filecontents*}{external_paper/tables/smape_main_table.csv}
Dataset,Frequency,LSTM_OURS,LSTM_OURS_SE,LSTM_MS,LSTM_MS_SE,LSTM_DR,LSTM_DR_SE,LSTM_RC,LSTM_RC_SE,CNN_OURS,CNN_OURS_SE,CNN_MS,CNN_MS_SE,CNN_DR,CNN_DR_SE,CNN_RC,CNN_RC_SE,Transformer_OURS,Transformer_OURS_SE,Transformer_MS,Transformer_MS_SE,Transformer_DR,Transformer_DR_SE,Transformer_RC,Transformer_RC_SE,S4_OURS,S4_OURS_SE,S4_MS,S4_MS_SE,S4_DR,S4_DR_SE,S4_RC,S4_RC_SE,ETS,ARIMA
M1,M,0.8838,0.1518,0.6001,0.0921,0.2027,0.0007,0.8644,0.1172,0.1914,0.0009,0.1945,0.0007,0.1974,0.0027,0.2308,0.0064,0.1953,0.0041,0.2104,0.0050,0.2126,0.0023,0.2335,0.0073,0.1954,0.0004,0.1975,0.0013,0.2077,0.0029,0.2315,0.0047,0.1693,0.1807
M1,Q,1.0103,0.1186,0.5547,0.0914,0.1508,0.0023,0.9659,0.1832,0.1410,0.0006,0.1369,0.0010,0.1431,0.0018,0.1479,0.0017,0.1456,0.0028,0.1505,0.0062,0.1615,0.0046,0.1574,0.0041,0.1456,0.0010,0.1425,0.0021,0.1533,0.0011,0.1596,0.0053,0.1624,0.1881
M1,Y,1.2673,0.2036,1.0504,0.2436,0.1622,0.0021,1.2674,0.2578,0.1601,0.0022,0.1490,0.0026,0.1504,0.0023,0.1565,0.0041,0.1690,0.0019,0.1773,0.0051,0.1761,0.0072,0.1616,0.0045,0.1626,0.0025,0.1612,0.0023,0.1785,0.0042,0.1674,0.0041,0.2188,0.1908
M3,O,0.4495,0.1494,0.2826,0.0485,0.0469,0.0013,0.3481,0.0925,0.0360,0.0003,0.0416,0.0003,0.0471,0.0013,0.0505,0.0023,0.0477,0.0015,0.0527,0.0031,0.0846,0.0267,0.0572,0.0064,0.0417,0.0008,0.0408,0.0030,0.0510,0.0043,0.0593,0.0036,0.0433,0.0438
M3,M,0.5901,0.1733,0.4287,0.0836,0.1583,0.0008,0.5309,0.1017,0.1486,0.0002,0.1510,0.0009,0.1510,0.0007,0.1887,0.0029,0.1542,0.0026,0.1605,0.0017,0.1663,0.0038,0.1904,0.0068,0.1488,0.0003,0.1487,0.0011,0.1513,0.0014,0.1854,0.0036,0.1627,0.1645
M3,Q,0.8036,0.1348,0.3693,0.0364,0.1073,0.0014,0.7295,0.1438,0.0986,0.0010,0.0985,0.0006,0.1039,0.0010,0.1099,0.0020,0.1047,0.0033,0.1122,0.0038,0.1214,0.0020,0.1147,0.0041,0.0981,0.0013,0.0920,0.0010,0.1115,0.0016,0.1156,0.0013,0.1014,0.1020
M3,Y,1.2251,0.1724,0.8595,0.1732,0.1720,0.0012,1.2075,0.2064,0.1658,0.0019,0.1573,0.0027,0.1508,0.0019,0.1677,0.0032,0.1624,0.0055,0.1688,0.0053,0.1753,0.0067,0.1687,0.0045,0.1689,0.0023,0.1547,0.0017,0.1680,0.0028,0.1726,0.0009,0.1971,0.1953
M4,H,0.8672,0.1765,0.6679,0.1069,0.2322,0.0136,0.8421,0.1295,0.2236,0.0019,0.2228,0.0030,0.2327,0.0104,0.3135,0.0094,0.2084,0.0064,0.2328,0.0174,0.2972,0.0094,0.3703,0.0181,0.2274,0.0025,0.1921,0.0014,0.1788,0.0067,0.3492,0.0291,0.2336,0.1426
M4,D,0.2972,0.1313,0.1540,0.0418,0.0319,0.0005,0.1878,0.0459,0.0302,0.0002,0.0303,0.0001,0.0304,0.0002,0.0321,0.0005,0.0332,0.0011,0.0309,0.0004,0.0334,0.0015,0.0370,0.0016,0.0308,0.0001,0.0301,0.0003,0.0318,0.0007,0.0371,0.0011,0.0280,0.0285
M4,W,0.4587,0.1497,0.2629,0.0539,0.0979,0.0010,0.3781,0.0832,0.0887,0.0004,0.0930,0.0007,0.0929,0.0020,0.0942,0.0017,0.0956,0.0017,0.0974,0.0016,0.1003,0.0029,0.1052,0.0054,0.0925,0.0013,0.0914,0.0017,0.0971,0.0031,0.1031,0.0033,0.0903,0.0878
M4,M,0.5515,0.1717,0.4187,0.0804,0.1445,0.0017,0.4866,0.0930,0.1397,0.0007,0.1402,0.0009,0.1397,0.0011,0.1597,0.0021,0.1370,0.0018,0.1428,0.0015,0.1533,0.0088,0.1785,0.0099,0.1357,0.0008,0.1401,0.0012,0.1397,0.0013,0.1726,0.0050,0.1321,0.1381
M4,Q,0.6446,0.1388,0.2833,0.0143,0.1199,0.0007,0.5502,0.1049,0.1124,0.0002,0.1117,0.0002,0.1117,0.0007,0.1220,0.0011,0.1238,0.0018,0.1237,0.0023,0.1291,0.0027,0.1327,0.0024,0.1152,0.0005,0.1174,0.0015,0.1176,0.0015,0.1273,0.0007,0.1045,0.1107
M4,Y,1.1435,0.1617,0.7597,0.1457,0.1561,0.0009,1.1286,0.2059,0.1503,0.0014,0.1426,0.0017,0.1424,0.0028,0.1524,0.0025,0.1466,0.0052,0.1528,0.0053,0.1602,0.0041,0.1473,0.0051,0.1555,0.0019,0.1421,0.0010,0.1545,0.0010,0.1585,0.0008,0.1546,0.1616
Tourism,M,1.0228,0.1849,0.7346,0.1128,0.3498,0.0088,1.0463,0.1921,0.3180,0.0045,0.3200,0.0033,0.3198,0.0056,0.4517,0.0214,0.3605,0.0030,0.3727,0.0055,0.3766,0.0071,0.4672,0.0291,0.3159,0.0029,0.3024,0.0007,0.3089,0.0049,0.4662,0.0143,0.2220,0.2206
Tourism,Q,0.9561,0.1394,0.4654,0.0205,0.2833,0.0052,0.9156,0.1928,0.2221,0.0020,0.2324,0.0031,0.2588,0.0059,0.2839,0.0038,0.2947,0.0037,0.2990,0.0042,0.3021,0.0039,0.2975,0.0120,0.2175,0.0029,0.2166,0.0034,0.2585,0.0096,0.2856,0.0049,0.1797,0.1785
Tourism,Y,1.3570,0.2463,1.0668,0.2368,0.2241,0.0045,1.1680,0.2732,0.2076,0.0030,0.2179,0.0057,0.2016,0.0031,0.2086,0.0035,0.1865,0.0029,0.1975,0.0020,0.2052,0.0089,0.2059,0.0043,0.2077,0.0036,0.1973,0.0012,0.2052,0.0015,0.2041,0.0008,0.2870,0.3030
\end{filecontents*}
\csvreader[]{external_paper/tables/smape_main_table.csv}
{Dataset=\dset,
Frequency=\frequency,
LSTM_OURS=\lstmours,
LSTM_OURS_SE=\lstmoursse,
LSTM_MS=\lstmms,
LSTM_MS_SE=\lstmmsse,
LSTM_DR=\lstmdr,
LSTM_DR_SE=\lstmdrse,
LSTM_RC=\lstmrc,
LSTM_RC_SE=\lstmrcse,
CNN_OURS=\cnnours,
CNN_OURS_SE=\cnnoursse,
CNN_MS=\cnnms,
CNN_MS_SE=\cnnmsse,
CNN_DR=\cnndr,
CNN_DR_SE=\cnndrse,
CNN_RC=\cnnrc,
CNN_RC_SE=\cnnrcse,
Transformer_OURS=\transours,
Transformer_OURS_SE=\transoursse,
Transformer_MS=\transms,
Transformer_MS_SE=\transmsse,
Transformer_DR=\transdr,
Transformer_DR_SE=\transdrse,
Transformer_RC=\transrc,
Transformer_RC_SE=\transrcse,
S4_OURS=\sfourours,
S4_OURS_SE=\sfouroursse,
S4_MS=\sfourms,
S4_MS_SE=\sfourmsse,
S4_DR=\sfourdr,
S4_DR_SE=\sfourdrse,
S4_RC=\sfourrc,
S4_RC_SE=\sfourrcse,
ETS=\ets,
ARIMA=\arima}
{%
\IfStrEq{\dset}{\prevdataset}{%
    \xdef\prevdataset{\dset}%
    \def\printdataset{}%
}{%
    \xdef\prevdataset{\dset}%
    \def\printdataset{\dset}%
}%

\printdataset & \frequency
& \lstmours & \lstmms & \lstmdr & \lstmrc 
& \cnnours & \cnnms & \cnndr & \cnnrc 
& \transours & \transms & \transdr & \transrc 
& \sfourours & \sfourms & \sfourdr & \sfourrc 
& \ets & \arima \\
& 
& \SE{\lstmoursse} & \SE{\lstmmsse} & \SE{\lstmdrse} & \SE{\lstmrcse} 
& \SE{\cnnoursse} & \SE{\cnnmsse} & \SE{\cnndrse} & \SE{\cnnrcse} 
& \SE{\transoursse} & \SE{\transmsse} & \SE{\transdrse} & \SE{\transrcse} 
& \SE{\sfouroursse} & \SE{\sfourmsse} & \SE{\sfourdrse} & \SE{\sfourrcse}
& \SE{-} & \SE{-} \\
\IfStrEq{\frequency}{Y}{%
    \IfStrEq{\dset}{tourism}{\\\bottomrule}{\\[-4pt]\hline\\[-3pt]}%
}{\hphantom{}}
}
\end{tabular}}
\end{table}

The \emph{symmetric mean absolute percentage error} is
\begin{equation}
    \mathrm{sMAPE} = \frac{1}{N \times T \times H } \sum_{i} \sum_{t} \sum_h \frac{2\,|y^{(i)}_{t+h}-\hat{y}^{(i)}_{t+h}|}{|y^{(i)}_{t+h}|+|\hat{y}^{(i)}_{t+h}|},
\label{eq:smape}
\end{equation}

\clearpage
\begin{table}[t!]
\caption{Empirical evaluation of point forecasts. Mean absolute scaled error (MASE, Eq.~\eqref{eq:mase}) with the \Naive baseline, averaged over 5 runs. Lower measurements are preferred; values below 1 beat the \Naive. The methods without standard deviation have deterministic solutions. For each encoder we compare four forecasting strategies: our \ourscomplete (\ours), multi-step (MS), direct (\DIR), and recurrent (\REC).}

\label{table:main_mase}
\centering
\scriptsize
\resizebox{\textwidth}{!}{
\begin{tabular}{ll|cccc|cccc|cccc|cccc|cc}
\toprule
& & \multicolumn{4}{c|}{\LSTM}
& \multicolumn{4}{c|}{\CNN}
& \multicolumn{4}{c|}{\Transformer}
& \multicolumn{4}{c|}{\Sfour}
& \multicolumn{2}{c}{StatsForecast} \\
& Freq 
& \multicolumn{1}{c}{\ours} & \multicolumn{1}{c}{MS} & \multicolumn{1}{c}{\DIR} & \multicolumn{1}{c|}{\REC} 
& \multicolumn{1}{c}{\ours} & \multicolumn{1}{c}{MS} & \multicolumn{1}{c}{\DIR} & \multicolumn{1}{c|}{\REC}
& \multicolumn{1}{c}{\ours} & \multicolumn{1}{c}{MS} & \multicolumn{1}{c}{\DIR} & \multicolumn{1}{c|}{\REC}
& \multicolumn{1}{c}{\ours} & \multicolumn{1}{c}{MS} & \multicolumn{1}{c}{\DIR} & \multicolumn{1}{c|}{\REC} 
& \ETS & \ARIMA \\
\midrule
\xdef\prevdataset{}
\begin{filecontents*}{external_paper/tables/mase_main_table.csv}
Dataset,Frequency,LSTM_OURS,LSTM_OURS_SE,LSTM_MS,LSTM_MS_SE,LSTM_DR,LSTM_DR_SE,LSTM_RC,LSTM_RC_SE,CNN_OURS,CNN_OURS_SE,CNN_MS,CNN_MS_SE,CNN_DR,CNN_DR_SE,CNN_RC,CNN_RC_SE,Transformer_OURS,Transformer_OURS_SE,Transformer_MS,Transformer_MS_SE,Transformer_DR,Transformer_DR_SE,Transformer_RC,Transformer_RC_SE,S4_OURS,S4_OURS_SE,S4_MS,S4_MS_SE,S4_DR,S4_DR_SE,S4_RC,S4_RC_SE,ETS,ARIMA
M1,M,0.8129,0.0062,0.8817,0.0054,0.9279,0.0045,0.9403,0.0168,0.8295,0.0048,0.9047,0.0080,0.9196,0.0217,1.0171,0.0439,0.8971,0.0231,0.9421,0.0097,0.9732,0.0225,1.0009,0.0101,0.9019,0.0032,0.8957,0.0097,0.9437,0.0282,1.0026,0.0143,0.7319,0.7660
M1,Q,0.9274,0.0268,0.9426,0.0140,0.9765,0.0262,0.9207,0.0211,0.9095,0.0047,0.9353,0.0136,0.9897,0.0204,1.2168,0.1247,1.0114,0.0771,1.0400,0.0244,1.0644,0.0465,0.9591,0.0228,0.9560,0.0201,0.9363,0.0144,1.0444,0.0143,0.9269,0.0204,0.8980,0.8899
M1,Y,0.8187,0.0209,0.7250,0.0253,0.7733,0.0274,0.9049,0.0833,0.7830,0.0132,0.8005,0.0391,0.6960,0.0475,0.7472,0.0430,0.6603,0.0319,0.8033,0.0853,0.8408,0.0841,0.7948,0.0561,0.7373,0.0180,0.8117,0.0378,0.7340,0.0509,0.8250,0.0534,0.7838,0.7534
M3,O,0.2018,0.0021,0.2106,0.0219,0.2080,0.0065,0.2241,0.0137,0.1552,0.0014,0.1859,0.0033,0.2074,0.0068,0.3381,0.0660,0.2191,0.0118,0.2490,0.0097,0.3886,0.1386,0.2517,0.0209,0.1786,0.0018,0.1754,0.0118,0.2250,0.0163,0.2634,0.0159,0.6915,0.7040
M3,M,0.6646,0.0049,0.6846,0.0077,0.7538,0.0068,0.8394,0.0257,0.6967,0.0012,0.7135,0.0032,0.7131,0.0034,1.0855,0.0458,0.7364,0.0182,0.7654,0.0109,0.7978,0.0207,0.9903,0.0668,0.7033,0.0022,0.7012,0.0053,0.7147,0.0040,0.9417,0.0254,0.8555,0.8622
M3,Q,0.7772,0.0097,0.7874,0.0118,0.8512,0.0120,0.8999,0.0217,0.7719,0.0090,0.7832,0.0091,0.8322,0.0147,1.0647,0.0942,0.8695,0.0523,0.9260,0.0376,0.9670,0.0166,0.9168,0.0434,0.7687,0.0090,0.7247,0.0081,0.8996,0.0044,0.9328,0.0118,0.9277,0.9257
M3,Y,0.8905,0.0141,0.7684,0.0093,0.8637,0.0104,0.9120,0.0521,0.8189,0.0118,0.7786,0.0093,0.7543,0.0143,0.9559,0.0705,0.8759,0.0725,0.8620,0.0455,0.8649,0.0300,0.8243,0.0249,0.8637,0.0119,0.7666,0.0105,0.8610,0.0106,0.8814,0.0082,0.9716,0.9868
M4,H,0.1617,0.0015,0.1703,0.0053,0.3738,0.0510,0.6322,0.0982,0.2715,0.0100,0.4464,0.0072,0.4665,0.0373,0.6923,0.0797,0.2363,0.0156,0.4631,0.0706,0.6826,0.0395,0.9682,0.0736,0.3018,0.0080,0.2829,0.0232,0.2617,0.0288,0.5793,0.0344,0.4148,0.1899
M4,D,0.7924,0.0081,0.7412,0.0130,0.8025,0.0195,0.8704,0.0309,0.7872,0.0095,0.7749,0.0071,0.7940,0.0109,0.8499,0.0537,0.8241,0.0296,0.7998,0.0201,0.8317,0.0295,0.9598,0.0594,0.7759,0.0076,0.7461,0.0072,0.8034,0.0187,0.9363,0.0377,0.9787,0.9892
M4,W,0.7441,0.0203,0.6783,0.0086,0.7613,0.0087,0.7483,0.0113,0.7249,0.0032,0.7357,0.0037,0.7323,0.0148,0.8055,0.0300,0.7847,0.0138,0.7858,0.0197,0.7907,0.0195,0.8308,0.0191,0.7272,0.0092,0.7404,0.0048,0.7746,0.0218,0.8350,0.0258,0.9487,0.9179
M4,M,0.6707,0.0027,0.6889,0.0111,0.7013,0.0077,0.7915,0.0164,0.6857,0.0045,0.6847,0.0041,0.6791,0.0057,0.9764,0.0621,0.6821,0.0122,0.7108,0.0126,0.7503,0.0463,0.9730,0.0712,0.6589,0.0025,0.6818,0.0046,0.6799,0.0087,0.9073,0.0321,0.8897,0.9006
M4,Q,0.7112,0.0083,0.7054,0.0088,0.7606,0.0044,0.7661,0.0110,0.7034,0.0016,0.7037,0.0031,0.7162,0.0056,0.8519,0.0267,0.8109,0.0299,0.7952,0.0165,0.8268,0.0166,0.8831,0.0349,0.7290,0.0047,0.7425,0.0105,0.7624,0.0082,0.8340,0.0126,0.8727,0.9168
M4,Y,0.7954,0.0109,0.7184,0.0061,0.7837,0.0051,0.8424,0.0503,0.7364,0.0077,0.7151,0.0070,0.7143,0.0155,0.8398,0.0558,0.7559,0.0597,0.7986,0.0576,0.7956,0.0204,0.7277,0.0254,0.7748,0.0086,0.7088,0.0058,0.7888,0.0049,0.8042,0.0061,0.8720,0.9138
Tourism,M,0.5549,0.0068,0.5803,0.0178,0.7457,0.0310,0.7938,0.0408,0.6228,0.0065,0.6564,0.0099,0.6554,0.0186,0.9876,0.0330,0.8165,0.0075,0.8057,0.0065,0.8230,0.0139,1.0525,0.1068,0.6349,0.0071,0.6108,0.0117,0.6114,0.0066,1.1097,0.0424,0.4376,0.5105
Tourism,Q,0.6449,0.0109,0.6428,0.0141,0.7976,0.0054,0.7594,0.0137,0.5982,0.0061,0.6236,0.0134,0.7131,0.0124,0.8791,0.0306,0.8221,0.0307,0.8556,0.0416,0.8282,0.0072,0.8397,0.0500,0.5810,0.0092,0.6212,0.0319,0.7454,0.0274,0.8607,0.0264,0.6232,0.6894
Tourism,Y,0.8995,0.0198,0.8550,0.0232,0.9295,0.0284,0.9524,0.0489,0.8570,0.0209,0.9025,0.0297,0.8281,0.0243,0.8323,0.0149,0.7907,0.0266,0.8730,0.0577,0.9087,0.0525,0.8869,0.0064,0.8973,0.0099,0.8785,0.0221,0.9466,0.0329,0.8508,0.0266,0.8804,1.0160
\end{filecontents*}
\csvreader[]{external_paper/tables/mase_main_table.csv}
{Dataset=\dset,
Frequency=\frequency,
LSTM_OURS=\lstmours,
LSTM_OURS_SE=\lstmoursse,
LSTM_MS=\lstmms,
LSTM_MS_SE=\lstmmsse,
LSTM_DR=\lstmdr,
LSTM_DR_SE=\lstmdrse,
LSTM_RC=\lstmrc,
LSTM_RC_SE=\lstmrcse,
CNN_OURS=\cnnours,
CNN_OURS_SE=\cnnoursse,
CNN_MS=\cnnms,
CNN_MS_SE=\cnnmsse,
CNN_DR=\cnndr,
CNN_DR_SE=\cnndrse,
CNN_RC=\cnnrc,
CNN_RC_SE=\cnnrcse,
Transformer_OURS=\transours,
Transformer_OURS_SE=\transoursse,
Transformer_MS=\transms,
Transformer_MS_SE=\transmsse,
Transformer_DR=\transdr,
Transformer_DR_SE=\transdrse,
Transformer_RC=\transrc,
Transformer_RC_SE=\transrcse,
S4_OURS=\sfourours,
S4_OURS_SE=\sfouroursse,
S4_MS=\sfourms,
S4_MS_SE=\sfourmsse,
S4_DR=\sfourdr,
S4_DR_SE=\sfourdrse,
S4_RC=\sfourrc,
S4_RC_SE=\sfourrcse,
ETS=\ets,
ARIMA=\arima}
{%
\IfStrEq{\dset}{\prevdataset}{%
    \xdef\prevdataset{\dset}%
    \def\printdataset{}%
}{%
    \xdef\prevdataset{\dset}%
    \def\printdataset{\dset}%
}%

\printdataset & \frequency
& \lstmours & \lstmms & \lstmdr & \lstmrc 
& \cnnours & \cnnms & \cnndr & \cnnrc 
& \transours & \transms & \transdr & \transrc 
& \sfourours & \sfourms & \sfourdr & \sfourrc 
& \ets & \arima \\
& 
& \SE{\lstmoursse} & \SE{\lstmmsse} & \SE{\lstmdrse} & \SE{\lstmrcse} 
& \SE{\cnnoursse} & \SE{\cnnmsse} & \SE{\cnndrse} & \SE{\cnnrcse} 
& \SE{\transoursse} & \SE{\transmsse} & \SE{\transdrse} & \SE{\transrcse} 
& \SE{\sfouroursse} & \SE{\sfourmsse} & \SE{\sfourdrse} & \SE{\sfourrcse}
& \SE{-} & \SE{-} \\
\IfStrEq{\frequency}{Y}{%
    \IfStrEq{\dset}{tourism}{\\\bottomrule}{\\[-4pt]\hline\\[-3pt]}%
}{\hphantom{}}
}
\end{tabular}}
\end{table}

The \emph{mean absolute scaled error}~\citep{hyndman2006another_look_measures} scales the absolute error by that of the \Naive forecast $\tilde{y}^{(i)}_{t+h}$, which repeats the last observation at the forecast creation date over the whole horizon:
\begin{equation}
    \mathrm{MASE} = \frac{\sum_{i} \sum_{t} \sum_h |y^{(i)}_{t+h}-\hat{y}^{(i)}_{t+h}|}{\sum_{i} \sum_{t} \sum_h |y^{(i)}_{t+h}-\tilde{y}^{(i)}_{t+h}|}.
\label{eq:mase}
\end{equation}
We scaled by the \Naive error in the same forecast windows, as in the relAvgRMSE, rather than by the seasonal error in-sample \Naive in M4's definition.

 \clearpage

\section{Universal Periodic Operator Approximation Theorem}
\label{appendix:extrapolation_bounds}

Theorem~\ref{theorem:universal_approximation} guaranties the approximation only in the training interval $[0,H]$. For periodic targets, a period that fits within the training interval is sufficient to extend the guaranty to all horizons $h \ge 0$.

\begin{theorem}[Universal Periodic Operator Approximation Theorem]
\label{theorem:periodic_approximation}
Assume the conditions of Theorem~\ref{theorem:universal_approximation}. Additionally, assume that for every history $\mathbf{y}_{:t} \in \mathcal{Y}$ the forecast function $f(\cdot \mid \mathbf{y}_{:t})$ extends to all $h \ge 0$ and is periodic with a period $J < H$,
\begin{equation}
    f(h + J \mid \mathbf{y}_{:t}) = f(h \mid \mathbf{y}_{:t}), \qquad h \ge 0,\quad \mathbf{y}_{:t} \in \mathcal{Y}.
\end{equation}
Then for any accuracy $\epsilon > 0$, there exists a neural network approximation $g_{\mathrm{per}}(h \mid \mathbf{y}_{:t}, \bw^{\star}) = \sum_{k=1}^{K} \btheta_k(\mathbf{y}_{:t}) \cdot \bphi_k(h \bmod J)$ such that, for every integer $m \ge 0$,
\begin{equation}
    \int_{mJ}^{(m+1)J} \left( f(h \mid \mathbf{y}_{:t}) - g_{\mathrm{per}}(h \mid \mathbf{y}_{:t}, \bw^{\star}) \right)^2 dh < \epsilon, \quad \forall \mathbf{y}_{:t} \in \mathcal{Y}.
\end{equation}
\end{theorem}

\begin{proof}
The proof has three steps.
\begin{enumerate}
    \item \textbf{One period approximation.} By Theorem~\ref{theorem:universal_approximation}, there exist neural basis functions $\{\bphi_k\}_{k=1}^{K}$ and coefficient mappings $\{\btheta_k\}_{k=1}^{K}$ such that $g(h \mid \mathbf{y}_{:t}, \bw^{\star}) = \sum_{k=1}^{K} \btheta_k(\mathbf{y}_{:t})\,\bphi_k(h)$ satisfies $\int_0^H (f - g)^2\,dh < \epsilon$ for all $\mathbf{y}_{:t} \in \mathcal{Y}$. Because $J < H$, the period $[0,J]$ is within the training interval and the integrand is nonnegative, so $\int_0^J (f - g)^2\,dh < \epsilon$.
    \item \textbf{Periodic extension.} For $h \in [0,J)$, $h \bmod J = h$, so $g_{\mathrm{per}}$ equals $g$ in the first period and $\int_0^J (f - g_{\mathrm{per}})^2\,dh < \epsilon$.
    \item \textbf{In every period.} Both $f$ and $g_{\mathrm{per}}$ are $J$-periodic in $h$. The substitution $h = mJ + u$ with $u \in [0,J)$ gives $\int_{mJ}^{(m+1)J} (f - g_{\mathrm{per}})^2\,dh = \int_0^J (f - g_{\mathrm{per}})^2\,du < \epsilon$ for every $m \ge 0$.
\end{enumerate}
\end{proof}

 \clearpage

\end{document}